\documentclass[journal]{IEEEtran}

\usepackage{cite}
\usepackage[pdftex]{graphicx}
\usepackage{xcolor}
\usepackage{tikz}
\usepackage{subfigure}
\usepackage{amsmath}
\usepackage{amsfonts}
\usepackage{bm}
\usepackage{mathtools}
\usepackage[super]{nth}
\usepackage[ruled, vlined, commentsnumbered, linesnumbered]{algorithm2e}
\usepackage{algorithmic}
\usepackage{array}
\usepackage{multirow}
\usepackage{url}
\usepackage{hyperref}
\usepackage{tabularx}
\usepackage{booktabs}

\definecolor{atomictangerine}{rgb}{1.0, 0.6, 0.4}
\definecolor{ballblue}{rgb}{0.13, 0.67, 0.8}
\definecolor{blue-violet}{rgb}{0.54, 0.17, 0.89}
\definecolor{ceruleanblue}{rgb}{0.16, 0.32, 0.75}
\definecolor{coolblack}{rgb}{0.0, 0.18, 0.39}
\definecolor{darkblue}{rgb}{0.0, 0.0, 0.55}
\definecolor{dogwoodrose}{rgb}{0.84, 0.09, 0.41}
\newcommand{\eg}{{\em e.g.}}           
\newcommand{\ie}{{\em i.e.}}           
\newcommand{\etc}{{\em etc.}}         

\newcommand{\HI} {\color{black}}
\newcommand{\JS} {\color{black}}
\newcommand{\RJS} {\color{black}}

\newcommand{\rev}[1]{\textcolor{black}{#1}}

\SetKwComment{tcp}{// }{}%
\newcommand{\nonl}{\renewcommand{\nl}{\let\nl\oldnl}}

\begin{document}
\title{A Plug-in Method for Representation Factorization in Connectionist Models}

\author{
Jee~Seok~Yoon, Myung-Cheol~Roh, and Heung-Il~Suk,~\IEEEmembership{Member,~IEEE}
\thanks{This work was partially done during J.S. Yoon's internship at Kakao Corp.
This study was supported by the Institute of Information \& Communications Technology Planning \& Evaluation (IITP) grant funded by the Korea Government (MSIT) (No. 2017-0-01779, A machine learning and statistical inference framework for explainable artificial intelligence), No. 2019-0-00079, Department of Artificial Intelligence (Korea University)), and Kakao Corp. (Development of Algorithms for Deep Learning-Based One-/Few-shot Learning).

J.S. Yoon is with the Department of Brain and Cognitive Engineering, Korea University, Seoul 02841, Korea (e-mail: wltjr1007@korea.ac.kr).

M.-C. Roh is with the Kakao Enterprise, Gyeonggi 13494 (e-mail: joshua.ai@kakaoenterprise.com).

H.-I. Suk is with the Department of Artificial Intelligence and the Department of Brain and Cognitive Engineering, Korea University, Seoul 02841 Korea (e-mail: hisuk@korea.ac.kr). (\textit{Corresponding author: Heung-Il Suk})}%
}

\markboth{IEEE TRANSACTIONS ON NEURAL NETWORKS AND LEARNING SYSTEMS}
{Yoon \MakeLowercase{\textit{et al.}}: \MakeUppercase{A Plug-in Method for Representation Factorization in Connectionist Models}}

\maketitle
\begin{abstract}
\nonl
In this paper, we focus on decomposing latent representations in generative adversarial networks or learned feature representations in deep autoencoders into semantically controllable factors in a semi-supervised manner, without modifying the original trained models. Particularly, we propose Factors Decomposer-Entangler Network (FDEN) that learns to decompose a latent representation into mutually independent factors.
Given a latent representation, the proposed framework draws a set of interpretable factors, each aligned to independent factors of variations by minimizing their total correlation in an information-theoretic means.
As a plug-in method, we have applied our proposed FDEN to the existing networks of Adversarially Learned Inference and Pioneer Network and performed computer vision tasks of image-to-image translation in semantic ways, \eg, changing styles while keeping the identity of a subject, and object classification in a few-shot learning scheme. We have also validated the effectiveness of the proposed method with various ablation studies in qualitative, quantitative, and statistical examination.
\end{abstract}


\IEEEpeerreviewmaketitle

\section{Introduction}
\IEEEPARstart{T}{he} advances in deep learning and its successes in various applications have been of significant interest for interpreting or {\RJS understanding the learned feature representations}. In particular, owing to a generic framework of deep generative adversarial learning, we have the tool of the Generative Adversarial Network (GAN) \cite{goodfellow2014generative} and its variants \cite{berthelot2017began,chen2016infogan,gulrajani2017improved}, to implicitly estimate the underlying data distribution in connection with a latent space. However, as the latent representation is highly entangled, it is still challenging to gain insights or interpret such latent representations in an observation space (\eg, an image).
A representation is generally considered disentangled when it can capture interpretable semantic information {\HI or} factors of the underlying variations {\HI in} the problem structure~\cite{bau2017network}.
Thus, the concept of disentangled representation is closely related to that of factorial representation~\cite{chen2018isolating,Kim20184153,schmidhuber1992learning}, which suggests that a unit of a disentangled representation should correspond to an independent factor of the observed data.
For example, {\HI there are different factors that describe a facial image, such as gender, baldness, smile, pose, identity.}
{\HI In this perspective, previous studies have also validated the effectiveness of disentangled representation} in various tasks such as few-shot learning~\cite{ridgeway2018learning,scott2018adapted,higgins2017darla,chen2018zero}, domain adaptation~\cite{zamir2018taskonomy,liu2018detach}, and image translation~\cite{gonzalez2018image,liu2018unified,chen2018isolating}.
While learning a disentangled representation is desirable, it does not imply that a (entangled) latent representation is less powerful or does not have any interpretability.
In fact, {\HI various methods that did not consider disentanglement~\cite{radford2015unsupervised,krizhevsky2012imagenet} achieved state-of-the-art performance in their respective domains}.
{\RJS Thus, when building deep models for any target tasks, it is desirable to achieve high performance and to have the learned feature representations interpretable or explainable by possibly making them disentangled. However, it is still very challenging to tackle those goals simultaneously, thus most of the researches in the literature focused on either of the problems. Notably, deep models that perform well on their respective tasks may not produce a disentangled and/or interpretable representation with respect to specific data generative factors. 
This motivated us to develop a novel `\emph{plug-in}' framework that helps disentangle learned feature representations of a deep model for better interpretation and explanation \emph{without modifying the original network architecture and trained model parameters as well as maintaining the performance on its original task}.}

\begin{figure}
	\centering
	\includegraphics[width=0.8\linewidth]{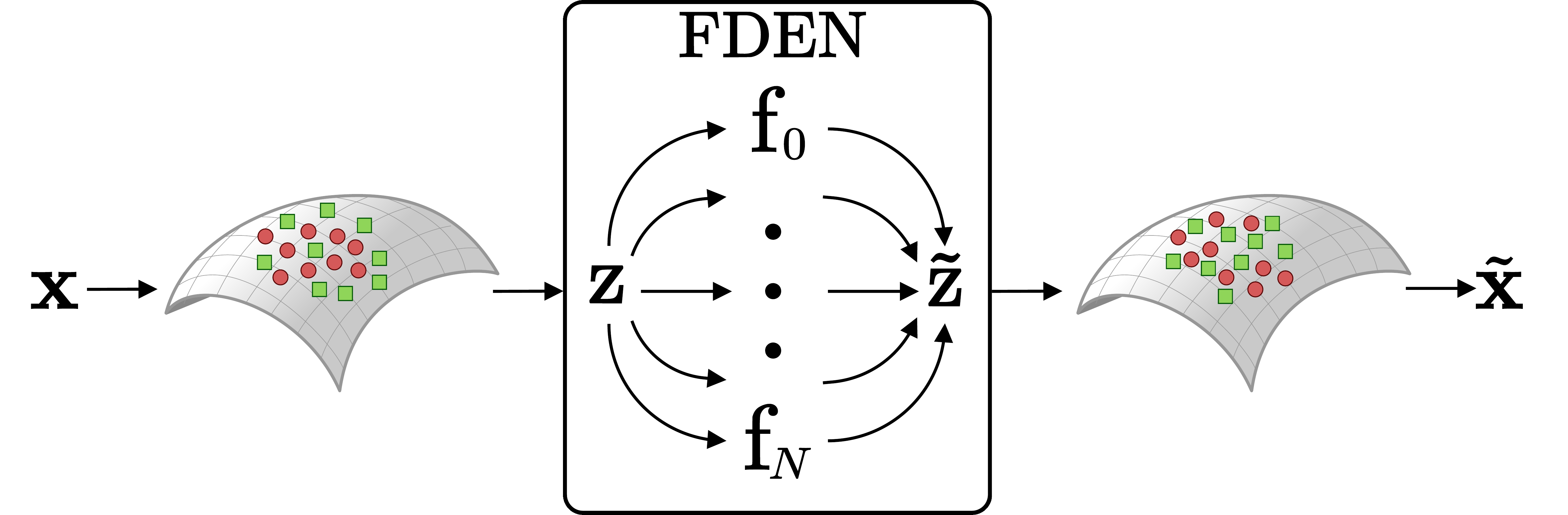}
	\caption{Overview of proposed framework. {\HI Factors Decomposer}-Entangler Network (FDEN) uses representation $\textbf{z}$ as input from \textit{fixed} pretrained model and outputs a reconstructed representation~$\tilde{\textbf{z}}$. In doing so, FDEN can factorize the representation into independent factors using information-theoretic approaches.}
	\label{overview_proposed}
\end{figure}

{\RJS Meanwhile, our proposed factorization module is applicable to decompose an entangled representation in any trained model into disentangled factors that could be used for other downstream tasks than it was originally trained for. 
For example, in our experiments, we have demonstrated to perform few-shot learning and image-to-image translation by taking a representation layer from a pretrained deep models, \ie, Adversarially Learned Inference (ALI) network~\cite{dumoulin2017} and Pioneer Network~\cite{heljakka2018pioneer}.}

In this study, given a pretrained deep model empowered with data generation such as GANs \cite{dumoulin2017,berthelot2017began} or Deep AutoEncoders (DAEs) \cite{heljakka2018pioneer,chen2018isolating}, we focus on decomposing the latent representations in GANs or learned feature representations in DAEs into semantically controllable factors in a semi-supervised manner, without modifying the original trained models. {\HI In particular, we devise Factors} Decomposer-Entangler Network (FDEN) that learns to decompose a representation into {\HI semantically} independent factors {\HI in a semi-supervised manner}. {\HI For a latent or feature representation vector}, the proposed network draws a set of interpretable factors, {\HI(some of which are derived in a supervised way when such information for input data is available)}, {\HI which are information-theoretically \textit{minimized} in mutual information}.
In addition, it can restore the independent factors back into its original representation, making FDEN an autoencoder-like architecture.
The {\RJS reason} behind the autoencoder-like architecture is to utilize the latent representation from a \textit{fixed} pretrained model rather than to develop and train a disentangled representation from scratch.
In doing so, we can focus our efforts solely on disentanglement with the benefit of the performance {\HI achieved by the pretrained model itself}.
{\HI Note that our method follows} a general consensus on a robust representation learning by (a) disentangling as many factors as possible, (b) maintaining maximum information in the original data~\cite{bengio2013representation}. 

{\RJS The motivation of our work is to take an information-rich, but entangled, representation and decompose it into interpretable factors.
This motivation may propose an important pathway since it is one of few works~[5] that tries to understand the actual interactions between or within representation layers.
A practical application of FDEN is a natural plug-in extension for well-trained models to be able to perform different tasks than it was originally designed to do so.
For example, we have taken a representation layer from a pretrained autoencoder and simultaneously performed few-shot learning and image style transfer.}
To evaluate our proposed framework, we perform qualitative, quantitative, and statistical examination of the factorized representation.
First, we measure the effectiveness of factorized representation in downstream tasks by performing image-to-image translation in conjunction with few-shot learning.
Then, we examine how each component of FDEN works toward creating a factorized representation with {\HI exhaustive} ablation studies and statistical analysis.
The main contributions of our study are as follows:

\begin{itemize}
	\item We propose a novel {\HI network, called {\HI Factors Decomposer}-Entangler Network (FDEN), that can be easily plugged in an existing network empowered with data generation.}
	\item {\RJS We} propose a novel approach for the \textit{minimization} of mutual information and total correlation with neural networks.
	\item {\HI Owing to the factorization property, our network can be used for image-to-image translation in semantic ways, \eg, changing styles while keeping the identity of a subject, and for classification tasks in a few-shot learning scheme.}
	\item Our study opens up the possibilities of extending {\RJS state-of-the-art generative and disentanglement models} to perform various tasks without modifying the weights so that it can maintain the performance of its original task.
\end{itemize}

\section{Related Works}
\subsection{Exploiting the Representation Vector}
There is a consensus~\cite{bengio2013representation,higgins2018towards,liu2018exploring} among many researchers that a robust approach to representation learning is through disentanglement.
To the best of our knowledge, previous research on disentangled representation has been focused on \textit{unsupervised} approaches to make each unit of a representation vector interpretable and independent of other units~\cite{higgins2017beta,Kim20184153}.
For example, Kim~\textit{et~al.}~\cite{Kim20184153} evaluate their representation on the classification performance of predicting which index of a representation corresponds to a factor of variation.
However, recent studies have pointed out flaws in unsupervised approaches to disentanglement and suggested exploring (semi-) supervised approaches to disentanglement~\cite{locatello2019challenging}.
To this end, Bau \textit{et~al.}~\cite{bau2019seeing, bau2017network} take a more direct and {\RJS semi-supervised approach to exploit the units of a representation}.
In particular, they propose ways to exploit the units of pretrained neural networks to independently turn on or off the factor of variations.
This is achieved by altering the value of the unit and analyzing the changes in the classification performance.
In a similar manner, our work approaches disentanglement through a semi-supervised factorial learning method.
However our work considers the representation as a whole rather than a unit basis.

\subsection{Deep Learning Based Independent Component Analysis}
Embedding or restoring independent components in a representation has been an on-going research topic in representation learning for decades~\cite{comon1994independent,jutten2003advances,higgins2017beta}.
{\RJS Recent approaches include autoencoder-based~\cite{bau2017network,zamir2018taskonomy,huang2019exploring}, and factor decomposition-based~\cite{Kim20184153,chen2018zero} methods that tries to infer interpretable components in a representation layer.}
There have been approaches to directly minimize the dependency between two random variations by means of adversarial learning~\cite{liu2018unified,liu2018exploring} and feature normalization~\cite{zhu2017unpaired}.
With the advances in GANs, models exploiting mutual information~\cite{belghazi2018,chen2016infogan} and their variants~\cite{ozair2019wasserstein,higgins2017beta} have been proposed.
These studies propose indirect approaches to independent component analysis and use the dual representation of mutual information to \textit{maximize} the mutual dependency between the data sample and its representation vector.
Several approaches based on directly minimizing the mutual information have been proposed; however they are inapplicable to neural networks~\cite{pham2004fast} and ignore the dual upper bound term (\textit{i.e.}, supremum term in~(\ref{DV_KL})).
In contrast to these works, we introduce a direct approach to minimizing the dependency between random variables applicable to most deep neural networks.
\begin{figure*}
	\centering
	\includegraphics[width=0.8125\linewidth]{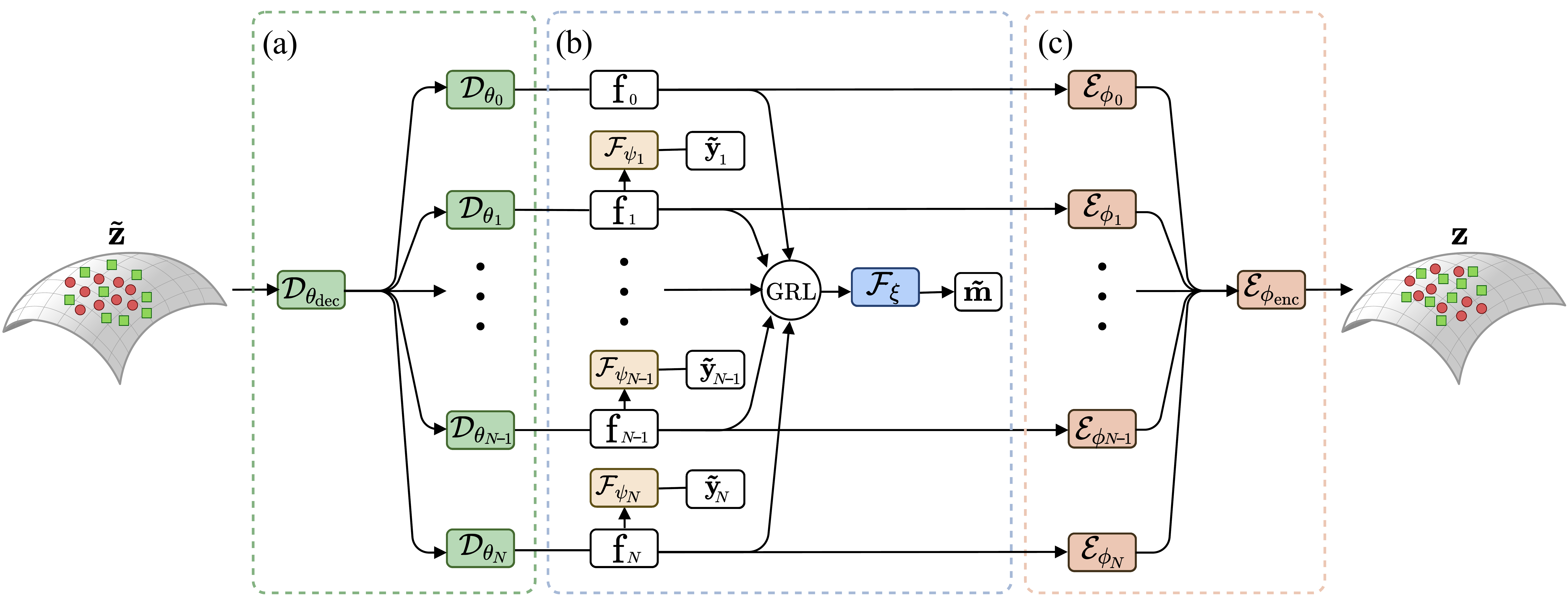}
	\caption{Overview of {\HI Factors Decomposer}-Entangler Network (FDEN). FDEN is divided into three modules: {\HI Decomposer} $\mathcal{D}$, Factorizer $\mathcal{F}$, and Entangler $\mathcal{E}$. The model is an autoencoder-like architecture that takes representation $\textbf{z}$ as the input and reconstructs its original representation $\tilde{\textbf{z}}$. (a) First, {\HI Decomposer} $\mathcal{D}$ takes a representation $\textbf{z}$ from a \textit{fixed} pretrained network as the input and decomposes it into a set of factors $\textbf{f}_i \left(\forall i \in N\right)$. (b) Next, Factorizer $\mathcal{F}$ uses an information theoretic way to maximize the independency of each factor. (c) Finally, Entangler $\mathcal{E}$ takes the factors and reconstructs their original representation~$\tilde{\textbf{z}}$.}
	\label{overview}
\end{figure*}

\section{Preliminary}
\subsection{Mutual Information}
{\HI In terms of an information theory, mutual information, which is a measure of the dependency between two random variables $X_0$ and $X_1$, can be formulated as the {\HI Kullback-Leibler (KL)} divergence as follows:}
\begin{equation}
\label{MI_KL}
I(X_0,X_1)=D_{KL}(\mathbb{P}_{X_0X_1}\rvert\lvert\mathbb{P}_{X_0} \otimes \mathbb{P}_{X_1})
\end{equation}
{\HI where $\mathbb{P}_{X_{0}X_{1}}$ denotes a joint probability distribution and $\mathbb{P}_{X_0} \otimes \mathbb{P}_{X_1}$ is the product of the marginal probability distributions $\mathbb{P}_{X_0}$ and $\mathbb{P}_{X_1}$.
{\RJS The \rev{intuitive understanding of} the KL-divergence in Eq.~\ref{MI_KL} is that the smaller the divergence between the joint and product of marginals, the more the independence between $X_0$ and $X_1$. In other words, if this divergence, \ie, mutual information, converges to zero, the two variables are \rev{mutually} independent to each other.}
Since it captures both linear and non-linear statistical dependencies between variables, mutual information is thought to be useful for measuring the true dependence} \cite{kinney2014equitability}. Therefore, we utilized mutual information {\HI in formulating our objective function as a means of} non-linearly decomposing a latent representation.
\subsection{Total Correlation}
Total correlation, or multi-information, is a variation of mutual information that can capture the dependency {\HI among} multiple random variables. For example, the total correlation {\HI among} a set of random variables $\left\{X_0, ..., X_N\right\}$ can be formulated as the KL-divergence between the joint probability $\mathbb{P}_{X_0...X_N}$ and the product of marginal probability $\mathbb{P}_{X_0} \otimes ... \otimes \mathbb{P}_{X_N}$:
\begin{equation}
I(X_0,...,X_N)=D_{KL}(\mathbb{P}_{X_0...X_N}\rvert\lvert\mathbb{P}_{X_0} \otimes ... \otimes \mathbb{P}_{X_N}).
\end{equation}
{\HI In Subsection~\ref{StatNet}, we discuss} how FDEN utilizes mutual information and total correlation.

\subsection{Donsker-Varadhan Representation of KL-divergence} Since mutual information and total correlation are intractable for continuous variables, we {\HI exploit} a dual representation~\cite{donsker1983asymptotic} for the KL-divergence computation:
\begin{equation}
\label{DV_KL}
{\RJS
D_{KL}(X\rvert\lvert Z)=\sup_{\xi}\mathbb{E}_{X}\left[T_{\xi}\right]-\log\left(\mathbb{E}_{Z}\left[\exp({T_{\xi})}\right]\right),
}
\end{equation}
where $T_{\xi} : \mathcal{X}\times\mathcal{Z}\rightarrow\mathbb{R}$ is a family of {\RJS functions parameterized $\xi$ by a neural network}. For full derivation of~(\ref{DV_KL}), readers are referred to~\cite{belghazi2018}.

\section{Factors Decomposer-Entangler Network}
FDEN is a novel framework that can be plugged into pretrained {\RJS connectionist models}, {\HI {\RJS especially but not limited to those} empowered with data generation (\eg, GANs) or reconstruction (\eg, DAEs)}, and factorize its {\HI latent or feature representation ${\bf z}$}.
In particular, the objective of FDEN is to decompose input representation ${\bf z}$ into independent and semantically interpretable factors without losing the original information in the latent or feature representation ${\bf z}$.
To achieve this aim, we {\HI compose an FDEN with} three modules (Fig.~\ref{overview}): {\HI Decomposer} $\mathcal{D}$, Factorizer $\mathcal{F}$, and Entangler $\mathcal{E}$.
Note that because FDEN uses a fixed pretrained network {\HI and deals with the latent or feature representation from the network, it allows factorizing the input representation for other new tasks while maintaining the network capacity or power for its original tasks {\RJS intact}}.

\subsection{Latent or Feature Representation}
The proposed FDEN has an autoencoder-like structure which uses {\HI a latent or feature} representation from a pretrained network {\RJS as input}.
{\HI For a pretrained network, we use networks capable of generating or encoding-decoding observable samples (\eg, an image).} 
In other words, we focus on {\HI deep} networks that {\HI find a latent representation from the input space and also reconstruct or generate a sample} given its latent representation.
Typical examples of these neural networks include bidirectional GANs~\cite{dumoulin2017,berthelot2017began}, autoencoders~\cite{vincent2010stacked,lu2013speech}, and invertible networks~\cite{behrmann2018invertible,jacobsen2018revnet}. {\RJS But it should be noted that it is not limited to those network but applicable to any connectionist models.}

\subsection{Decomposer-Entangler}
The {\HI Decomposer}-Entangler network (Fig.~\ref{overview}~(a) and (c)) is an autoencoder-like architecture that uses {\HI representation~$\textbf{z}$ as input} and reconstructs its original representation~$\tilde{\textbf{z}}$.
Particularly, {\HI Decomposer}~$\mathcal{D}$ takes {\HI a representation~$\textbf{z}$ as input} and decodes it with a global decoder network~$\mathcal{D}_{\theta_{\text{dec}}}$.
Next, the decoded representation $\textbf{z}_\text{dec}$ is {\HI decomposed} into a set of factors, {\HI each of which uses} a local decoder network, {\HI \eg, $\textbf{f}_{i}$~$\left(=\mathcal{D}_{\theta_{i}}\left(\textbf{z}_\text{dec}\right), \forall i\in N\right)$}.
Entangler~$\mathcal{E}$ takes factors~$\textbf{f}_{i} \left(\forall i\in N\right)$ into their corresponding streams {\HI $\mathcal{E}_{\phi_{i}}(\textbf{f}_{i}), \left(\forall i \in N\right)$}.
These streams are then concatenated on the channel axis and fed into the {\HI global} encoder $\mathcal{E}_{\phi_{\text{enc}}}$ to reconstruct the original representation $\tilde{\textbf{z}}\left(=\mathcal{E}_{\phi_{\text{enc}}}\left(\mathcal{E}_{\phi_{0}}\left(\textbf{f}_{0}\right) \oplus ... \oplus\mathcal{E}_{\phi_{N}}\left(\textbf{f}_{N}\right)\right)\right)$, {\RJS where $\oplus$ is a concatenation operator}.
Since the objective of the {\HI Decomposer}-Entangler network is to reconstruct the original representation {\RJS hopefully without any information loss in the procedural steps}, we introduce the $\ell_2$ reconstruction objective function $\mathcal{L}_{R}$. {\RJS When concerning the architecture of a pretrained network on which we conduct representation factorization}, because a sample $\textbf{x}$ and its representation $\textbf{z}$ may or may not be bijective, we include a regularizer to the reconstruction objective function {\HI as follows}:
\begin{equation}
\mathcal{L}_{R}=\lvert\lvert \textbf{z} - \tilde{\textbf{z}} \rvert\rvert^{2}_{2}+ \lambda \lvert\lvert \textbf{x} - \tilde{\textbf{x}} \rvert\rvert^{2}_{2},
\label{eq:lr_loss}
\end{equation}
where $\lambda$ is a constant weight term for the regularizer. Note that a \textit{fixed} pretrained network uses input $\tilde{\textbf{z}}$ to reconstruct its data $\tilde{\textbf{x}}$ (Fig.~\ref{overview_proposed}). {\RJS For connectionist models, if there is no sample reconstruction module is available this regularizer can be ignored.}

At this point, representation $\textbf{z}$ is merely decomposed and reassembled into $\tilde{\textbf{z}}$ (for {\HI an} ablation study on FDEN trained with only $\mathcal{L}_R$ objective function, refer to Subsection~\ref{remove_statnet}).
Although these factors contain information {\HI in ${\bf z}$}, they are not aligned to specific {\HI factors} of variation.
In other words, the factors are not independent, nor do they carry any distinguishable information.
Thus, we introduce a module, called Factorizer, to give information on these factors {\RJS in a semi-supervised manner as described} in the following subsection.

\subsection{Factorizer}
Factorizer $\mathcal{F}$ uses an information-theoretic measure to make the factors independent and obtain distinguishable information.
The general idea is to minimize the total correlation {\HI among} all factors ({\HI via} \textit{Statisticians Network}) while giving them {\RJS optionally} relevant information using a set of classifiers (\textit{Alignment Network}). 

\begin{figure*}
	\centering
	\subfigure[Schematic overview of image-to-image translation scenario]{\includegraphics[width=0.495\linewidth]{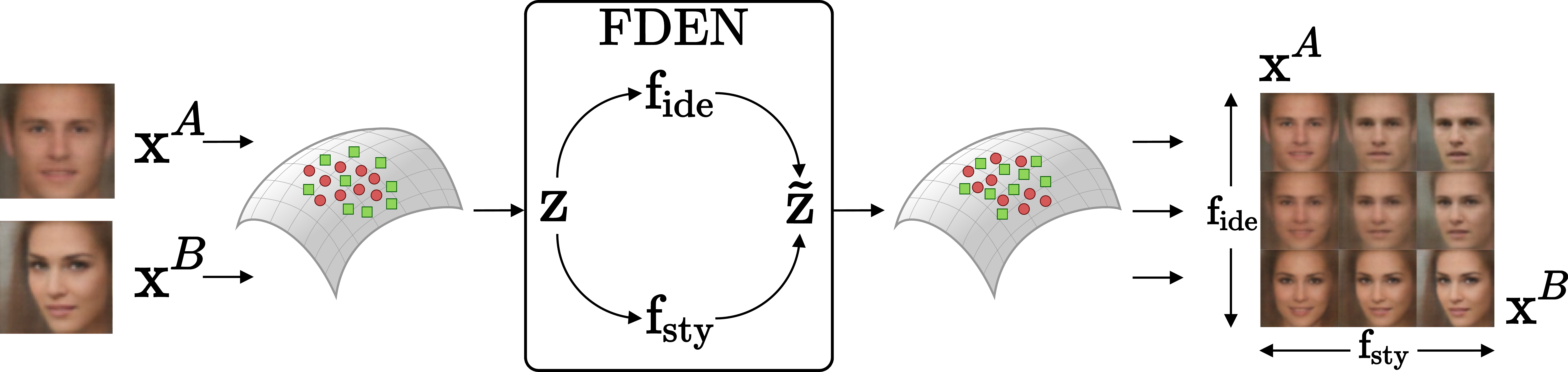}}
	\hfill
	\subfigure[FDEN in an image-to-image translation scenario]{\includegraphics[width=0.495\linewidth]{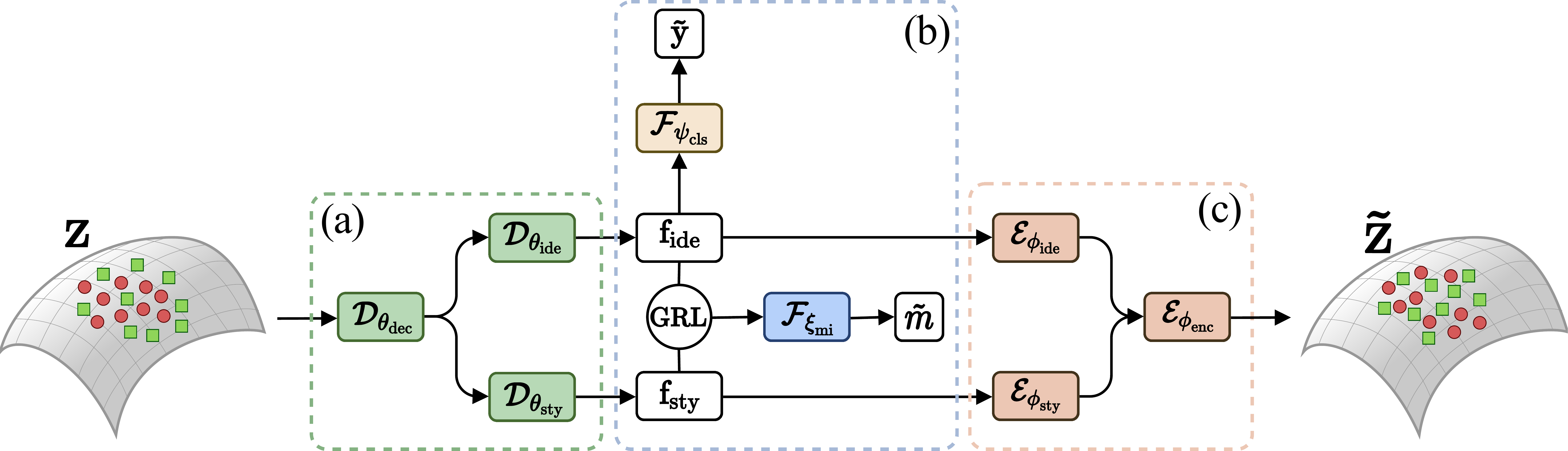}}
	\caption{
		FDEN in an image-to-image translation scenario.
		First, FDEN takes a latent representation $\textbf{z}$ as the input and decomposes it into an identity factor~$\textbf{f}_{\text{ide}}$ and a style factor~$\textbf{f}_{\text{sty}}$.
		Then, latent representation $\tilde{\textbf{z}}$ is reconstructed by linearly interpolating the factors of various representations (\textit{e.g.} $\tilde{\textbf{f}}^{AB}=\alpha\textbf{f}^{A}+\left(1-\alpha\right)\textbf{f}^{B}$).
	}
	\label{i2i_diagram}
\end{figure*}
\subsubsection{\label{StatNet}Statisticians Network}
The first component of Factorizer, Statisticians Network $\mathcal{F}_{\xi}$, estimates the total correlation {\HI among factors in a one-versus-all scheme}.
Our objective is to minimize the total correlation {\HI among} factors $\textbf{f}_{i} \left(\forall i \in N\right)$ so that they are {\RJS mutually independent to each other.}
We follow~\cite{belghazi2018} (\textit{i.e.}, Eq. (\ref{DV_KL})) to estimate the total correlation {\HI among} factors:
\begin{equation}
\label{mine_eq}
\mathcal{L}_{M}=\sup_{\xi}\mathbb{E}_{\mathbb{P}_{0,...,N}}\left[\mathcal{F}_{\xi}\right]-\log\left(\mathbb{E}_{\mathbb{P}_{0} \otimes...\otimes \mathbb{P}_{N}}\left[\exp\left(\mathcal{F}_{\xi}\right)\right]\right)
\end{equation}
where $\mathcal{F}_{\xi}$ is the Statisticians Network, $\mathbb{P}_{0,...,N}$ is the joint distribution of all factors (\textit{i.e.}, $\left(\textbf{f}_0,...,\textbf{f}_N\right) \sim \mathbb{P}_{0,...,N}$), and $\mathbb{P}_{0} \otimes...\otimes \mathbb{P}_{N}$ is the product of the marginal distributions of all factors.
We simplify the marginal distribution by taking $\textbf{f}_\text{0}\sim\mathbb{P}_0$ from the joint distribution $\left(\textbf{f}_0,...,\textbf{f}_N\right) \sim \mathbb{P}_{0,...,N}$ and $\bar{\textbf{f}}_i\sim\mathbb{P}_i \left(1<i<N\right)$ from the joint distribution shuffled \textit{i.i.d.} by the batch axis for each factor, \textit{i.e.} $\left(\textbf{f}_\text{0},\bar{\textbf{f}}_1,...,\textbf{f}_N \right),...,\left(\textbf{f}_\text{0},...,\bar{\textbf{f}}_{N-1}, \textbf{f}_N \right),\left(\textbf{f}_\text{0},...,\bar{\textbf{f}}_N \right) \sim \mathbb{P}_{0,...,N}$.
{\HI Although the latent representation is factorized into independent factors, from a semantic point of view, the decomposed factors are not necessarily {\RJS and intuitively} interpretable yet. In this regard, we further consider minimal networks that help factors to be mapped to the human understandable factor of variations in a supervised manner, {\RJS when such factors are available}.}

\subsubsection{Alignment Networks}
The Alignment Network is designed to link each factor to {\HI one of the human labeled factors (\eg, attributes) in a supervised manner. Concretely, there is a set of classifiers $\mathcal{F}_{\psi_{i}} \left(1<i<N\right)$ that identifies whether an input sample for latent representation ${\bf z}$ has the target factor {\RJS or attribute} information. This supervised learning implicitly guides each factor to be aligned with one of the factor labels.} 
Statisticians Network makes the factors independent to each other. Therefore, when one factor $\textbf{f}_i$ has information on a factor of variation, \eg, for {\HI gender}, the other factors, {\HI \ie, $\textbf{f}_{j\neq i}$, will have other independent information, {\RJS \eg, age, sunglasses}.
However, the existence of a significant number of factors that possibly make diverse variations in samples makes it unsuitable to consider the human labeled attributes only.
In this regard, we further consider another independent factor dedicated for other potential factors, not specified in human labels. This unspecified factor ${\bf f}_{0}$ is trained in an \textit{unsupervised} way, only being involved in total correlation minimization. 
To jointly train the Alignment Networks {\HI except for ${\bf f}_{0}$}, we define the supervised loss function as follows:
\begin{equation}
\mathcal{L}_{C_i}=CE(y_i,\hat{y}_i),
\mathcal{L}_C=\frac{1}{N-1}\sum_{i=1}^{N}{\mathcal{L}_{C_i}},
\end{equation}
{\RJS where the objective function is a cross-entropy function, and $\hat{y}_i=\mathcal{F}_{\psi_i}(\textbf{f}_i)$.}

It should be noted that this Alignment Network is capable of ensuring alignment between factors and human labeled attributes, because Statisticians Network causes the factors to be independent via total correlation minimization.
Further,} reconstruction loss $\mathcal{L}_{R}$ in Eq. (\ref{eq:lr_loss}) ensures that {\RJS the decomposed factors have no or minimal information loss}.

{\RJS In this sense, conceptually, the \textit{Factorizer} $\mathcal{F}$ is a pseudo-surjective\footnote{Pseudo- since the relationship is inferred.} function that maps $\textbf{z}\Rightarrow{}\textbf{f}_{i}~\left(\forall~i>0\right)$.
		This relationship allows for an interesting property that an input sample will have a pseudo-bijective relationship with a set of factors, \ie, $\textbf{x}\Leftrightarrow{}\left\{\textbf{f}_0,...,\textbf{f}_N\right\}$, regardless of the (non-)~bijective nature of functions mapping $\textbf{x}\xrightarrow{}\textbf{z}$ or $\textbf{z}\xrightarrow{}\textbf{x}$.
        The intuition behind pseudo-bijective relationship is that any input sample can be decomposed into a set of factors, and any combinations of factors can be reassembled to produce a sample in the original input space. Thus, one of natural applications of FDEN is style transfer in computer vision where we can change the values of a decomposed factor and replace it with the original factor to reconstruct an image with different style.}

\begin{figure*}
	\centering
	\includegraphics[width=0.8\linewidth]{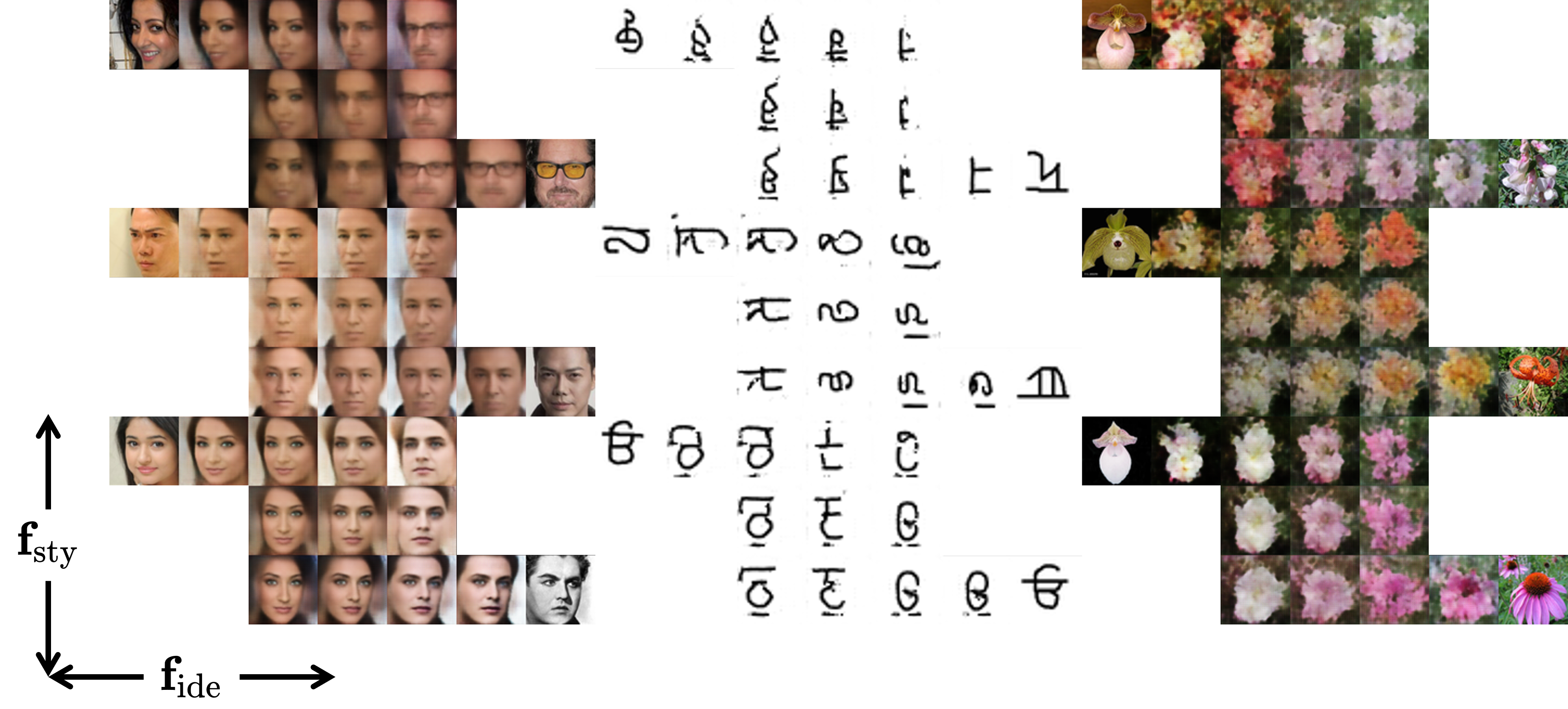}
	\caption{Results of image-to-image translation for the MS-Celeb-1M, Omniglot, and Oxford Flower datasets.
		For each dataset, images on the first and the last column are the input images to be translated.
		Images on the second and sixth columns are ALI's original reconstruction.
		Images in the middle are results of reconstruction with interpolated identity and style factors of the input images. Additional results are in the Supplementary Chapter A.
	}
	\label{i2i_trans}
\end{figure*}
\begin{figure*}
	\centering
	\subfigure[MS-Celeb-1M (Different celebrities)]{\includegraphics[width=0.425\linewidth]{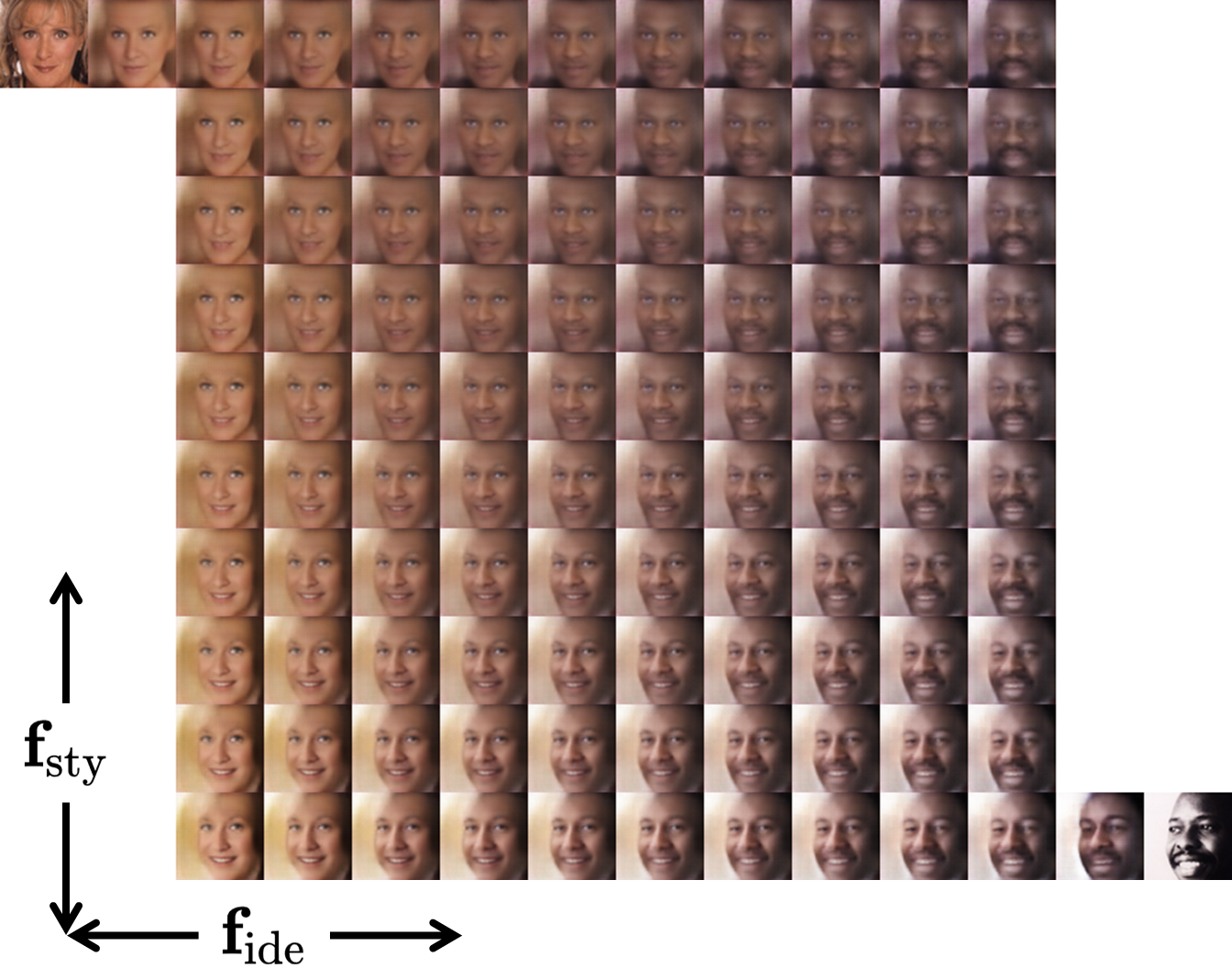}}
	\hfill
	\subfigure[Oxford Flower (Different flowers)]{\includegraphics[width=0.425\linewidth]{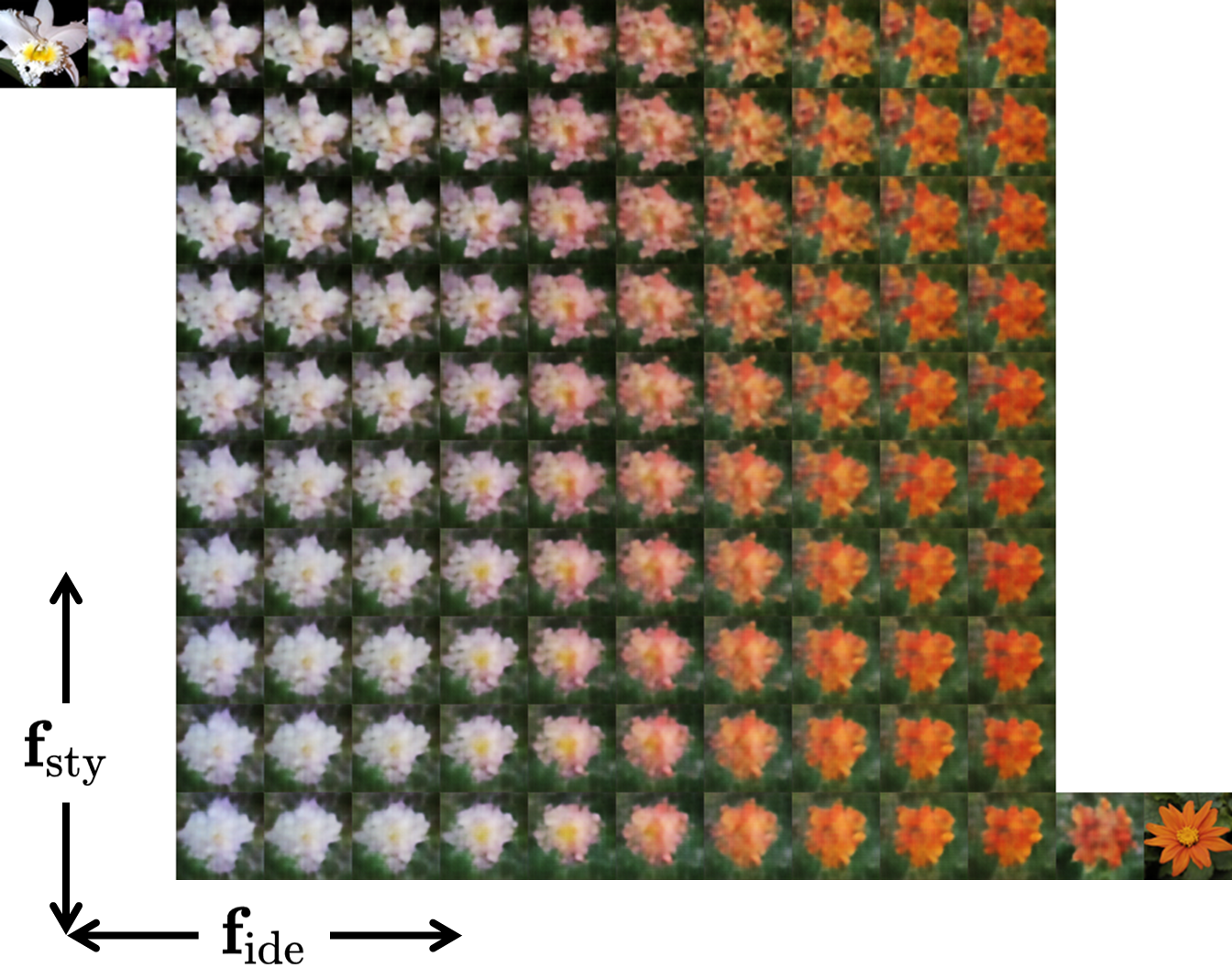}}
	\hfill
	\subfigure[MS-Celeb-1M (Same celebrity)]{\includegraphics[width=0.425\linewidth]{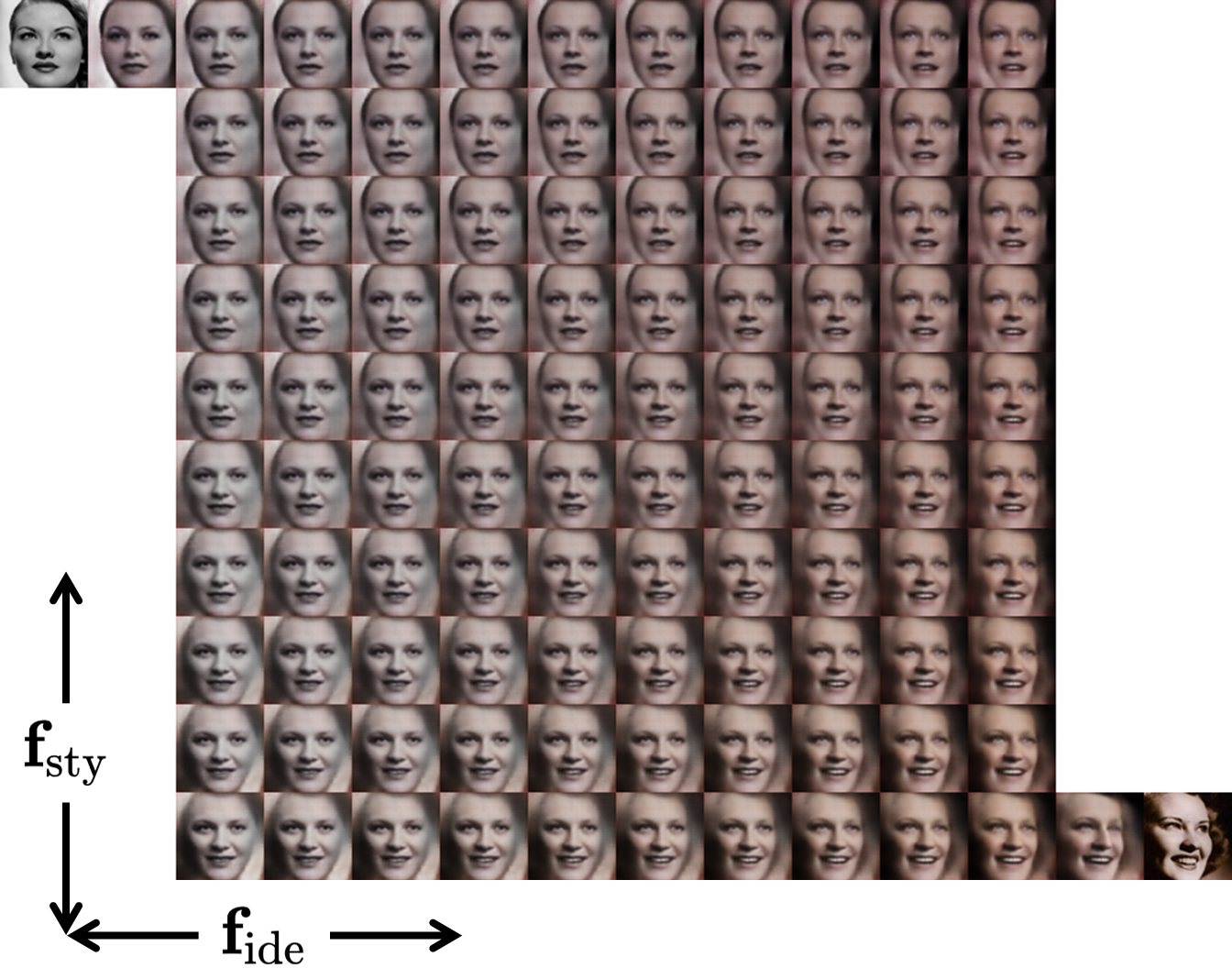}}
	\hfill
	\subfigure[Oxford Flower (Same flower)]{\includegraphics[width=0.425\linewidth]{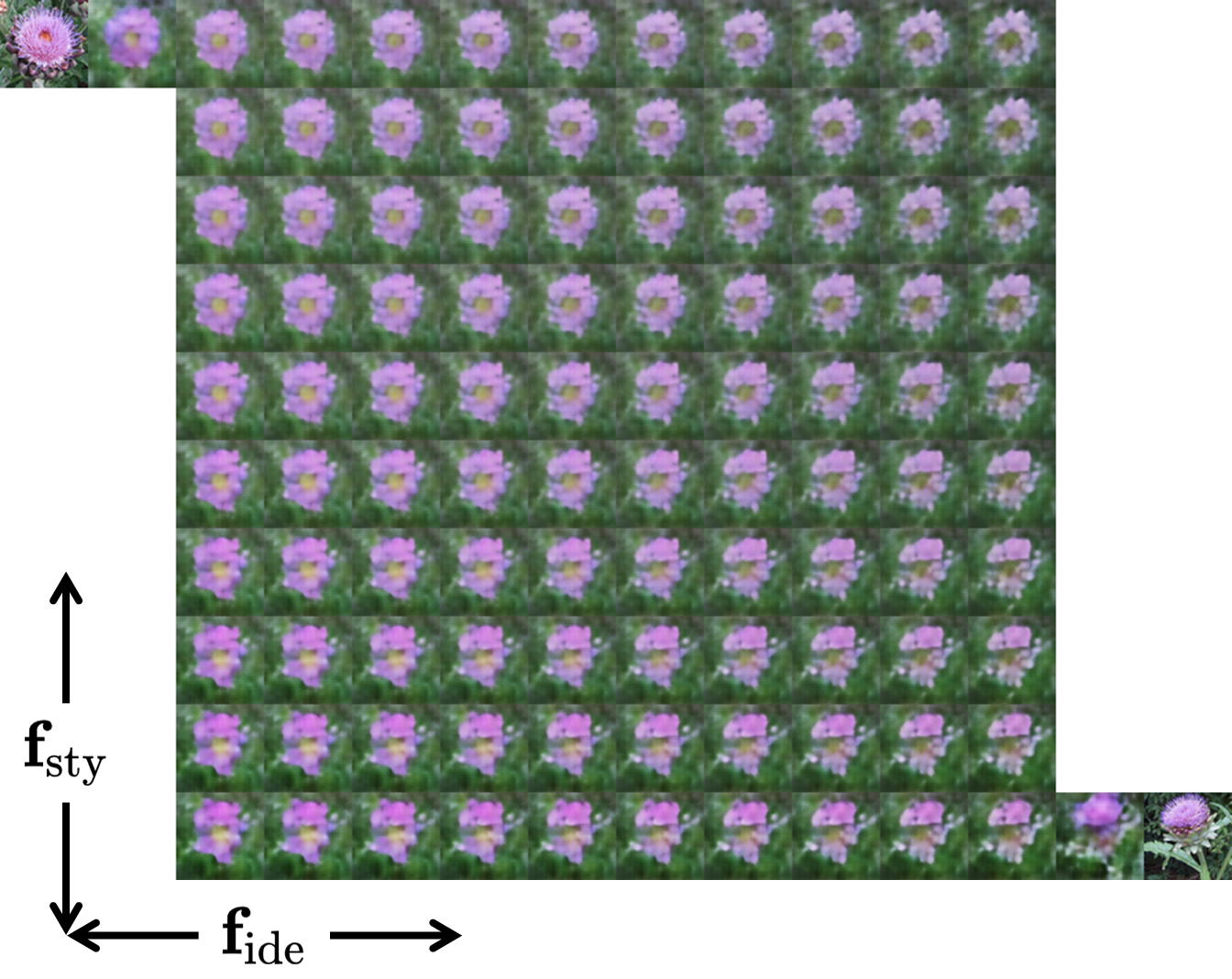}}
	\caption{
		Results of image-to-image translation for the MS-Celeb-1M and Oxford Flower datasets with fine interpolation between identity and style factors.
		(a, b) Translation is performed on images with different identities.
		(c, d) Translation is performed on images with the same identity.
	}
	\label{i2i_samediff}
\end{figure*}
\subsection{Learning}
We define the overall objective function for FDEN {\RJS as follows}:
\begin{equation}
\mathcal{L} = \alpha\mathcal{L}_{R}+\beta\mathcal{L}_{C}-\gamma\mathcal{L}_{M},	
\end{equation}
where $\alpha,\beta$, and $\gamma$ {\HI are the coefficients to weight different loss {\RJS terms}, and the negative} $\mathcal{L}_{M}$ is {\HI due to the maximization of~(\ref{mine_eq}) for its supremum term. Since we need to \textit{minimize} our objective and the dependency among factors}, we introduce a workaround using a Gradient Reversal Layer~\cite{ganin2016domain} in the following subsection.

\subsubsection{Gradient Reversal Layer (GRL)}
Note that $\mathcal{L}_{M}$ needs to be \textit{maximized} to successfully estimate the dual representation of the KL-divergence, but our aim is to \textit{minimize} the dependency among factors.
Thus, we add a GRL~\cite{ganin2016domain} before the first layer of {\HI Statisticians} Network.
In essence, {\HI the} GRL multiplies the gradients by a negative constant during backpropagation {\RJS only}.
With {\HI the} GRL in place, the Statisticians Network $\mathcal{F}_{\xi}$ will maximize $\mathcal{L}_M$ to estimate the total correlation; however, the rest of the network will be guided toward the minimization of mutual information (for details on the effectiveness of GRL against other approach, refer to Subsection~\ref{grl_ablate}).

\subsubsection{Adaptive Gradient Clipping}
Since $\mathcal{L}_{M}$ is unbounded, its gradients can overwhelm the gradients of other objective functions when left {\HI uncontrolled}.
To mitigate this problem, we apply an adaptive gradient clipping~\cite{belghazi2018}:
\begin{equation}
g_{a}=\min{\left(\lvert\lvert g_{u}\rvert\rvert,\lvert\lvert g_{m}\rvert\rvert\right)}\frac{g_{m}}{\lvert\lvert g_{m}\rvert\rvert},
\end{equation}
where $g_{a}$ is the adapted gradients, $g_{u}:=\frac{\partial\left(\mathcal{L}_{R}+\mathcal{L}_{C}\right)}{\partial\theta}$, and $g_{m}:=+\frac{\partial\mathcal{L}_{M}}{\partial\theta}$ (positive due to GRL). $g_{a}$ is the gradients over $\theta$ because $\mathcal{L}_{M}$ backpropagates only through $\theta$ and $\xi$.
\section{Experiments}
In this section, we perform various {\HI experiments} to {\RJS justify and evaluate the power of} FDEN.
Our objective here is to demonstrate that each module of FDEN is effective at decomposing a latent representation into independent factors.
{\RJS First, we evaluate the effectiveness of factors by performing various downstream tasks. 
Next, we analyze individual units of factors to verify if a representation is indeed reasonably factorized.
Finally, we perform ablation studies to evaluate the effectiveness of each module of FDEN in factorizing a representation\footnote{Code available at \url{https://github.com/wltjr1007/FDEN}}.}

\subsection{Datasets}
We evaluate {\HI the proposed FDEN} on datasets in various domains: Omniglot (character), MS-Celeb-1M (facial with identity), CelebA (facial with attributes), Mini-ImageNet (natural), {\RJS Oxford Flower (floral), and dSprites (2D shapes)} datasets.
\subsubsection{Omniglot} The Omniglot~\cite{lake2015human} dataset consists of 1,623 characters from 50 alphabets, where each character is drawn by 20 different people via Amazon's Mechanical Turk. 
Following~\cite{vinyals2016matching,snell2017prototypical}, we partitioned the dataset into 1,200 characters for training and remaining 423 for testing.
Also following~\cite{vinyals2016matching,snell2017prototypical}, we augmented the dataset by rotating 90, 180, 270 degrees, where each rotation is treated as a new character (\textit{i.e.}, 4,800 characters for the training dataset and 1,692 characters for the testing dataset).
\subsubsection{MS-Celeb-1M Low-shot} The MS-Celeb-1M~\cite{guo2016ms} low-shot dataset consists of facial images of 21,000 celebrities. This dataset is partitioned (by~\cite{guo2016ms}) into 20,000 celebrities for training and 1,000 celebrities for testing. There are average 58 images per celebrity in the training dataset (total of 1,155,175 images), and 5 images per celebrity in the test dataset (total of 5,000 images).
\subsubsection{CelebA} The CelebA~\cite{liu2015faceattributes} dataset consists of 202,599 celebrity facial images with 40 binary attributes, such as eye-glasses, bangs, smile.
The dataset is partitioned (by~\cite{liu2015faceattributes}) into 162,770 images for training, 19,867 images for validation, and 19,962 images for testing.
\subsubsection{Mini-ImageNet} Mini-ImageNet is a partition of the ImageNet dataset created by~\cite{ravi2016optimization} for few-shot learning. It consists of 100 classes from ImageNet with 600 images per class, and~\cite{ravi2016optimization} have split the dataset it into 64, 16, and 20 classes for training, validation, testing, respectively.
\subsubsection{Oxford Flower} The Oxford Flower~\cite{nilsback2008automated} dataset consists of images of 102 flower species, with 40 to 258 per flower species. We have split the dataset by randomly selecting 82 flower species for training and 20 flower species for testing.
{\RJS \subsubsection{dSprites} The dSprites dataset~\cite{dsprites17} is a collection of 2D shape images specifically designed for evaluating disentanglement. It consists of 737,280 grayscale images with 6 ground truth factor of variations (1 object color, 3 shapes, 6 scales, 40 orientations, 32 x position, 32 y position). Following~\cite{locatello2019challenging}, we do not use a separate train and test split since some disentanglement scores require interventions on the ground truths latent factors.}

\subsection{Implementation Details}
\subsubsection{Pretrained Networks}
For the pretrained network, we utilize Adversarially Learned Inference~(ALI)~\cite{dumoulin2017} and Pioneer Network~\cite{heljakka2018pioneer}.

ALI is a bidirectional GAN that simultaneously learns a generation network and an inference network.
We chose ALI for its simplicity in implementation and its ability to create powerful latent representation.
For MS-Celeb-1M, Mini-ImageNet, and Oxford dataset, we replicated the model designed in \cite{dumoulin2017} for the CelebA dataset.
For the Omniglot dataset, we replicated the model designed in \cite{dumoulin2017} for the SVHN dataset.

Pioneer Network~\cite{heljakka2018pioneer} is a progressively growing autoencoder capable of achieving high quality reconstructions.
We have chosen Pioneer Network also for its state-of-the-art reconstruction performance.
Apart from various GANs, Pioneer Network created one of the highest quality reconstructions we have found.
We use the pretrained model for CelebA-128 publicly available online\footnote{\url{https://github.com/AaltoVision/pioneer}} by~\cite{heljakka2018pioneer}.

\subsubsection{Factors Decomposer-Entangler Network}
FDEN consists of {\HI Decomposer}, Statisticians Network, Alignment Network, and Entangler, which are fully connected layers parameterized by $\theta, \xi, \psi$, and $\phi$, respectively.
For the sake of simplicity and model complexity, we kept each module to 3 or 4 fully connected layers with dropout, batch normalization, and a leaky ReLu activation. 

For details of hyperparameters, readers are referred to the Supplementary Chapter B.

\label{remove_statnet}
\subsection{Downstream Task}

\subsubsection{Image-to-Image Translation}
\label{subsection_i2i}
The objective of this experiment is to demonstrate the effectiveness of FDEN in decomposing and reconstructing a latent representation.
Given representations of two samples, $\textbf{z}^A$ and $\textbf{z}^B$, we perform image-to-image translation by linearly interpolating their identity factors, $\textbf{f}_{\text{ide}}^A$ and $\textbf{f}_{\text{ide}}^B$, with style factors of different images, $\textbf{f}_{\text{sty}}^A$ and $\textbf{f}_{\text{sty}}^B$ (Fig.~\ref{i2i_diagram}).
For example, $\tilde{\textbf{f}}^{AB}=\alpha\textbf{f}^{A}+\left(1-\alpha\right)\textbf{f}^{B}$.
Without modifying the weights of the invertible networks, we reconstruct a translated image with $\tilde{\textbf{z}}^{AB}\sim\left(\tilde{\textbf{f}}^{AB}_{\text{ide}},\tilde{\textbf{f}}^{AB}_{\text{sty}}\right)$.
For image-to-image translation, we evaluate our results with the Omniglot, MS-Celeb-1M, and Oxford Flower datasets using pretrained ALI (Fig.~\ref{i2i_trans} and Fig.~\ref{i2i_samediff}) .

\begin{figure*}
	\centering
	\includegraphics[width=0.9\linewidth]{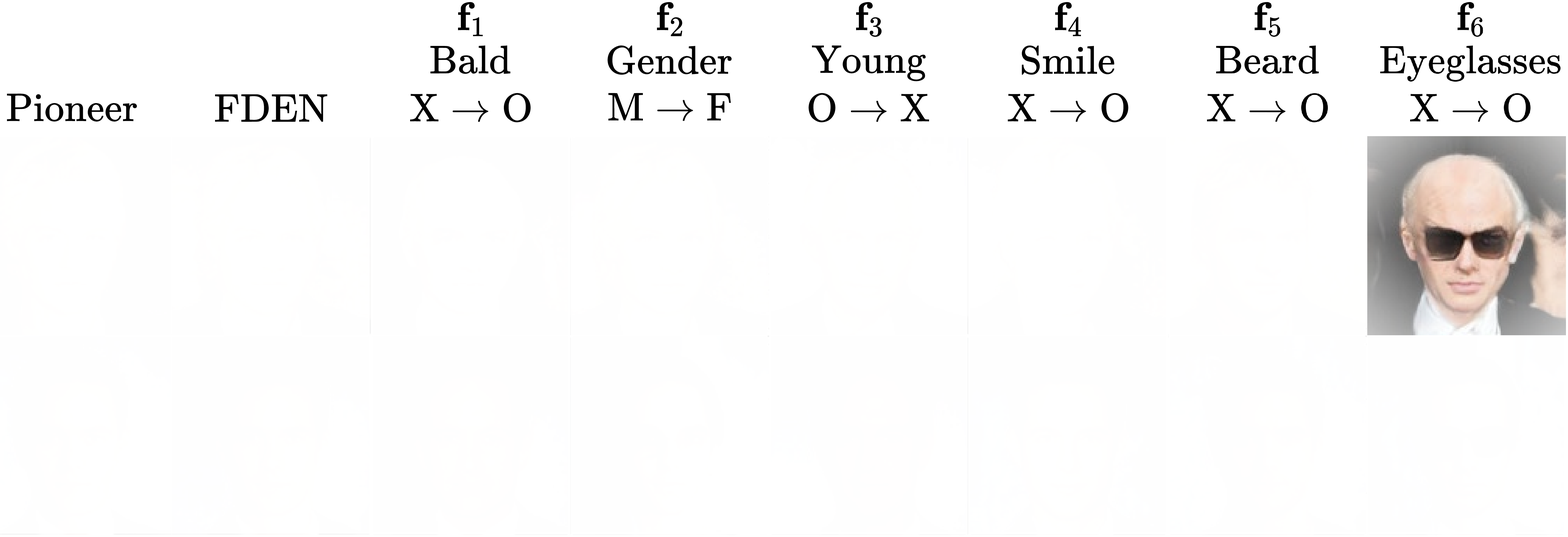}
	\caption{Results of style transfer for the CelebA-128 dataset with N=7 factors (where $\textbf{f}_{0}$ is the style factor).
	Images in the {\RJS first and second} columns are reconstructed images from Pioneer Network~\cite{heljakka2018pioneer} and FDEN, respectively.
	The following images are reconstructed images with one attribute opposite to the input image (\eg, \nth{1} row {\RJS $\textbf{f}_3$}: ``not bald'' transferred to ``bald''; \nth{2} row $\textbf{f}_3$: ``young'' transferred to ``not young'').
	The original attributes of both input images are: ``not bald'', ``male'', ``young'', ``without smile'', ``without beard'', ``without eyeglasses'' (note that the \nth{1} row image is annotated as ``with goatee, but without beard'').
	}
	\label{style_trans}
\end{figure*}
Our results show that identity-relevant features are clearly aligned with identity factors.
For example, the first MS-Celeb-1M images from Fig.~\ref{i2i_trans} depict clear interpolation between a woman and a man row-wise.
Since we only factorize a representation into two factors, style factor $\textbf{f}_{\text{sty}}$ carries all non-relevant information for identity.
Thus, during interpolation between factors, we see multiple factors changing together, such as changes in the rotation and brightness of the face and background.
Although it is hard to distinguish which factor of variation changes during interpolating factors of the Omniglot and Oxford Flower datasets, we notice that each step of interpolation results in a partially interpretable change. 
These observations indicate that FDEN can decompose a latent representation into independent factors.

Furthermore, comparing the reconstructed images from ALI (\nth{1} row \nth{2} column, \nth{6} row \nth{3} column) and FDEN(\nth{1} row \nth{3} column, \nth{3} row \nth{5} column), we observe that they are significantly similar.
This shows that FDEN can indeed be plugged into a pretrained network without reducing its performance on its original downstream task (additional hi-resolution results are available in the Supplementary Chapter A). 

{\JS To verify the independence between the identity and style factors more clearly, we perform a fine interpolation between identity and style factors with the same or different identities~(Fig.~\ref{i2i_samediff}).
The interpolation between style, \ie, row-wise interpolation, shows only the style related factor of variations change.
Similarly, interpolation between identity, \ie, column-wise interpolation, indicates that identity factors are changed only with different identities.}
{\JS

\subsubsection{Style Transfer}
To verify the {\RJS DV (Donsker-Varadhan)} representation of total correlation with multiple variables, we perform style transfer with human labeled attributes (Fig.~\ref{style_trans}).
For style transfer, FDEN is trained with the CelebA-128 dataset with multiple factors, where each factor is aligned to an attribute (except $\textbf{f}_0$).
Style transfer using attributes is performed similar to image-to-image translation, where the factor to be transferred is replaced by the mean factor of the opposite attribute.
For example, to transfer ``not bald'' attribute, \ie, $\textbf{f}_1$ in Fig.~\ref{style_trans}, to ``bald'' attribute, $\textbf{f}_1$ is replaced by the mean of $\textbf{f}_1$ from all bald celebrities while the rest of the factors remain.

The results of style transfer with FDEN confirms a clear transfer of attributes; however, in this process, other independent factors also change unintentionally.
For example, ``eyeglasses'' attribute ($\textbf{f}_6$) is accompanied with changes in ``bald'' attribute ($\textbf{f}_1$) for the first example in Fig.~\ref{style_trans}.
{\RJS We presume that} this is because of the inevitable gap in the bounds of the DV representation~\cite{mcallester2018formal}.
Since the DV representation is an estimate of mutual information, the more the factors, the larger the errors in the estimate.
Furthermore, we have adopted a linear interpolation approach by replacing the original factor with the mean factor of all samples with the opposite attribute; however, the factor vectors may not lie on a linear space.
We will discuss this further in the Discussion section.
}
\begin{table}
	\centering
	\caption{$C$-way $K$-shot learning accuracy.
		\textsc{\textbf{FDENf}} is FDEN trained with fixed pretrained network and \textsc{\textbf{FDENe}} is FDEN trained end-to-end with pretrained network.
		\textsc{\textbf{MLP}} is the baseline experiment with MLP classifier using only representation~$\textbf{z}$.}
	\label{fewshot_acc}
	\begin{tabular}{@{}cccccc@{}}
		\toprule
		& \multicolumn{3}{c}{\textbf{Omniglot}} & \multicolumn{2}{c}{\textbf{Mini-ImageNet}} \\
		\multicolumn{1}{l}{} & \multicolumn{2}{c}{\textbf{5-way}} & \textbf{20-way} & \multicolumn{2}{c}{\textbf{5-way}} \\
		& 1-shot & 5-shot & 1-shot & 1-shot & 5-shot \\ \midrule
		\textsc{\textbf{MatchNet}}~\cite{vinyals2016matching} & 98.1\% & 98.9\% & 93.8\% & 43.5\% & 55.3\% \\
		\textsc{\textbf{ProtoNet}}~\cite{snell2017prototypical} & \textbf{98.8\%} & \textbf{99.7\%} & \textbf{96.0\%} & 49.4\% & \textbf{68.2\%} \\
		\textsc{\textbf{FDENe}} & 91.1\% & 99.0\% & 90.7\% & \textbf{49.4\%} & 61.4\% \\ \midrule
		\textsc{\textbf{MLP}} & 80.3\% & 89.8\% & 65.2\% & 26.3\% & 37.2\% \\
		\textsc{\textbf{FDENf}} & \textbf{88.3\%} & \textbf{95.4\%} & \textbf{82.6\%} & \textbf{43.9\%} & \textbf{48.6\%} \\ \bottomrule
	\end{tabular}
\end{table}
\subsubsection{Few-shot Learning}

Several approaches for evaluating a representation have been proposed, most notably the \textit{disentanglement scores}~\cite{locatello2019challenging}.
{\RJS We} have referenced on some {\RJS classifier-based} disentanglement scores, such as FactorVAE~\cite{Kim20184153} and BetaVAE~\cite{higgins2017beta} scores, and found that the few-shot learning setup has a setting significantly similar to these scores. 
Therefore, we chose the few-shot learning performance as {\RJS a downstream task} to evaluate how much the factors are independent of each other.
For this experiment, Alignment Network exploits an episodic learning scheme \cite{vinyals2016matching} suitable for few-shot learning scenario.
Each episode consists of randomly sampled $C$ unique classes, $K$ support samples per class, and a query sample from one of the $C$ classes.
Given $C\times K$ support samples, the objective of the few-shot learning is to predict which of the $C$ unique classes does the query sample belong to.
In the few-shot learning literature, this setup is generally called the $C$-way, $K$-shot learning.

Here, we formally define the settings of episodic learning similar to that of~\cite{vinyals2016matching}.
First, we define episode $E$ as the distribution over all possible labels $L$, where a label set $L\sim E$ contains batches of randomly chosen $C$ unique classes.
Next, we define $S\sim L$ as the support set with $k$ data-label pairs $\left({\mathbf x},y\right)^k$, and $Q\sim L$ as the batches of a single data-label pair.
The objective of episodic learning is to match a query data-label pair with the support data-label pair of the same label.
Thus, we formulate the objective function of episodic learning as follows:
\begin{equation}
\label{fewshot_LC}
\mathcal{L}_{C}=\mathbb{E}_{L \sim E}\left[\mathbb{E}_{S \sim L,Q \sim L}\left[\sum_{\left({\mathbf x},y\right)\in Q}{\log{P\left(y\rvert {\mathbf x},S\right)}}\right]\right],
\end{equation}
where $\mathcal{L}_{C}$ is the cross-entropy objective function between predictions $\tilde{y}$~$\left(=P\left(y\rvert {\mathbf x},S\right)\right)$ and ground truths $y$.
{\RJS Each episode in an episodic learning scheme can be thought of as a mini few-shot task since it subsamples a few classes and data samples every episode. Thus, the training environment of the episodic learning scheme naturally generalizes to the test environment.

For the few-shot learning down-stream task, we have trained FDEN in an end-to-end manner with the episodic learning objective function~\ref{fewshot_LC} for the \textit{Alignment Network}. We denote \textsc{FDENf} as FDEN trained with fixed and pretrained $\mathbf{z}$, and \textsc{FDENe} as FDEN trained without fixing the pretrained $\mathbf{z}$. Since the comparison works~\cite{vinyals2016matching, snell2017prototypical} and FDEN shared the same episodic learning scheme and the dataset splits, the scores reported in Table~\ref{fewshot_acc}. are acquired from the corresponding papers.}

We evaluate FDEN on few-shot learning to demonstrate that the decomposed identity factor $\textbf{f}_\text{ide}$ is successful in containing the identity information of the observed data.
Thus, we validate our results on two different domains of data with varying complexities --- Omniglot and Mini-ImageNet --- and compare our results with studies that use the episodic learning scheme (\cite{vinyals2016matching, snell2017prototypical},~Table~\ref{fewshot_acc}).
One property of FDEN is that it learns to exploit only the latent space.
In other words, FDEN does not have any information on the input data except for a pretrained model's representation of it.
Thus, our baseline (denoted as \textsc{MLP}) for this experiment is the few-shot learning performance using only representation $\textbf{z}$ with an MLP classifier with the same structure as that of the FDEN's Alignment Network.
We have shown our results with the pretrained network fixed (denoted as \textsc{FDENf}) and end-to-end learning by fine-tuning both FDEN and the pretrained network (denoted as \textsc{FDENe}).
Note that we share the same weights for image-to-image translation experiments in Subsection~\ref{subsection_i2i} and for few-shot learning experiments.
We evaluate our results on 1,000 episodes with unseen samples for all experiments.

The results of \textsc{FDENf} and image-to-image translation indicate that the identity factors and style factors indeed contain information relevant to their factor of variation.
As for the results on end-to-end learning (\textit{i.e.} \textsc{FDENe}), the few-shot learning performance significantly improves compared to \textsc{FDENf}, but the quality of image-to-image translation slightly degrades due to the changes in weights of the pretrained model.
Although our results on end-to-end experiments are inferior when compared with other methods, {\RJS it should be emphasized that our FDEN was trained with networks originally designed for other tasks, rather than few-shot learning, it is reasonable why the performance of FDEN is lower than that of the few-shot oriented networks.} 

\begin{table}
	\centering
	\caption{{\RJS Comparison of disentanglement scores with competing \rev{methods in the literature on the} dSprites dataset. \textsc{\textbf{FDENu}} is FDEN trained in an unsupervised manner, and \textsc{\textbf{FDENs}} is FDEN trained in a supervised manner. (BVM: $\beta$-VAE Metric, FVM: FactorVAE Metric, DCI: Disentanglement \rev{(D), Completeness (C) and Informativeness (I)}, MIG: Mutual Information GAP)}}
	\label{quant_eval}
    \begin{tabular}{@{}ccccccc@{}}
		\toprule
                 & \textbf{BVM} & \textbf{FVM} & \textbf{MIG}    & \textbf{D} & \textbf{C} & \textbf{I} \\ \midrule
    \textbf{$\beta$-VAE}~\cite{higgins2017beta}      & 0.8476  & 0.6540    & 0.1059 & 0.1561          & 0.1697          & 0.3987       \\
    \textbf{FactorVAE}~\cite{Kim20184153}    & 0.8564  & 0.6918    & 0.1371 & 0.2144          & 0.2628          & 0.3896       \\
    \textbf{DIP-VAE}~\cite{Kumar2018VariationalIO}      & 0.8356  & 0.6436    & 0.1025 & 0.1248          & 0.1184          & 0.3705       \\
    \textbf{$\beta$-TCVAE}~\cite{chen2018isolating}    & 0.8472  & 0.7450    & 0.1050 & 0.1602          & 0.1589          & 0.3968       \\
    \textbf{AnnealVAE}~\cite{locatello2019challenging}  & 0.8384  & 0.7406    & 0.2593 & 0.3283          & 0.3893          & 0.2887       \\
    \textbf{IDGAN}~\cite{lee2020highfidelity}        & \textbf{0.8852}  & 0.7766    & 0.2311 & \textbf{0.4332}          & 0.4761          & \textbf{0.5201}       \\
    \textsc{\textbf{FDENu}} & 0.8325  & \textbf{0.7923}    & \textbf{0.4234} & 0.4211          & \textbf{0.4635}          & 0.4912       \\ \midrule
    \textsc{\textbf{FDENs}}   & 0.8823  & 0.8624    & 0.5992 & 0.5462          & 0.6201          & 0.7235      \\
    \bottomrule
    
    \end{tabular}
\end{table}

\subsection{Analysis}
\begin{figure}
	\centering
	\subfigure[Identity factors]{\includegraphics[width=0.49\linewidth]{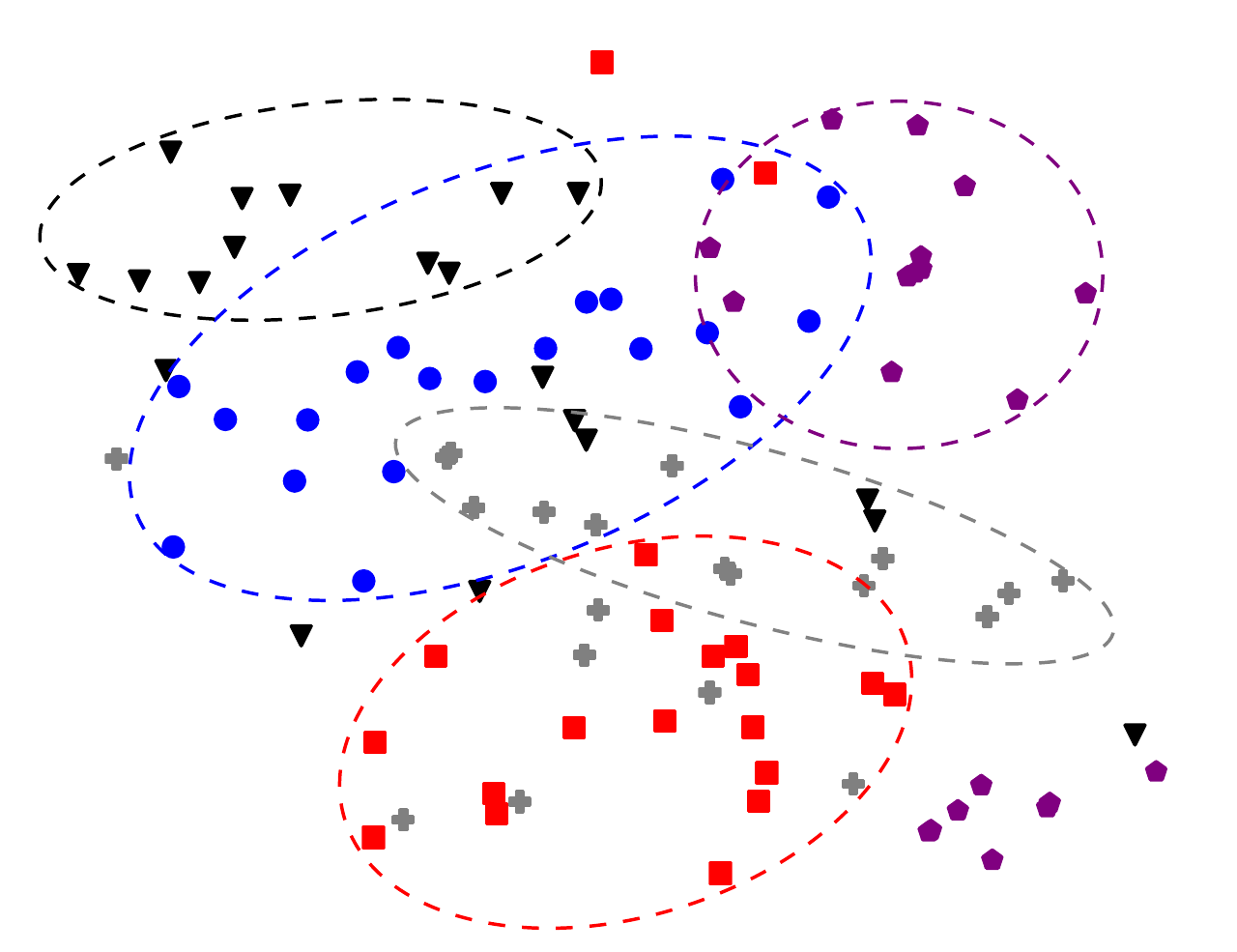}}
	\hfill
	\subfigure[Style factors]{\includegraphics[width=0.49\linewidth]{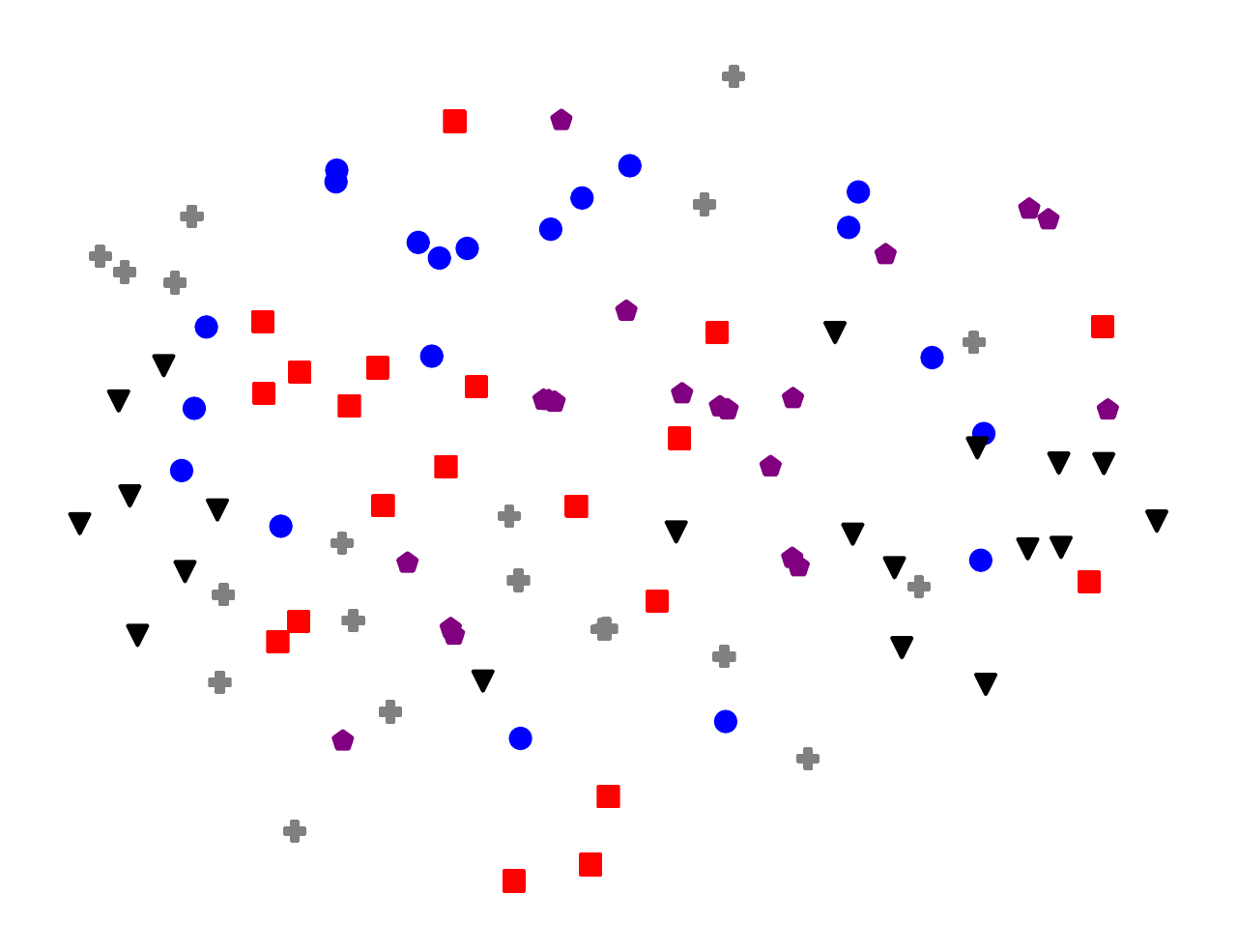}}
	
	\caption{t-SNE scatter plot of factors from 5-way 1-shot Omniglot model.
		As shown by the dotted lines in (a), the identity factors are clearly clustered when compared with style factors in (b).
		Each plot consists of 5 unique classes with 20 samples per class (best viewed in color).
	}
	\label{z_tsne}
\end{figure}

{\RJS
\subsubsection{Disentanglement Score}
    {\RJS To demonstrate that FDEN is able to \rev{decompose} factors in an unsupervised manner, we have performed an \rev{additional} experiment \rev{on the dSprites dataset, for which full attributes are provided, thus directly comparable among methods}, by removing the Alignment Network (\ie, classifier $\mathcal{F}_\psi$ and loss function $\mathcal{L}_{C}$) and compared the disentanglement scores with baseline ($\beta$-VAE, FactorVAE, $\beta$-TCVAE, and DIP-VAE) and state-of-the-art (AnnealVAE, IDGAN) works.
		
    For the pretrained network, we use the pretrained $\beta$-VAE (reported in Table~\ref{quant_eval}) which is publicly \rev{available} \footnote{\label{disentanglement_lib}\url{https://github.com/google-research/disentanglement_lib}, weights of all competing works except IDGAN were acquired from this URL (model number = 0: $\beta$-VAE, 300: FactorVAE, 600: DIP-VAE, 1200: $\beta$-TCVAE, 1500: AnnealVAE). IDGAN was trained with the default seed and settings provided at \url{https://github.com/1Konny/idgan}} by~\cite{locatello2019challenging}.
    Since the factors $\left\{\mathbf{f}_0,...,\mathbf{f}_4\right\}$ are separated into different streams, we have concatenated the factors into a single vector $\mathbf{f}_{\text{concat}}$ and randomly permuted the index once before evaluation \rev{to remove the possibility of exploiting grouped elements in a long vector for classification}.
    Also, for a fair comparison, the concatenated vector has the same dimension of 10 (\ie, $\mathbf{f}_\text{concat}\in\mathbb{R}^{10}$) as the competing works.
    For each of the scores, we have randomly selected 10,000 training samples and 5,000 test \rev{samples,} and evaluated the scores using the same seed and settings as~\cite{locatello2019challenging}.}
}

\subsubsection{t-SNE} To further analyze our results, we draw t-SNE scatter plots with factors from a 5-way 1-shot Omniglot model (Fig.~\ref{z_tsne}).
The t-SNE plot for identity factors shows apparent clusters between samples of the same class, whereas the style factors show no visible clusters.
This observation suggests that identity factors are indeed aligned to identity information (in this case, a letter).
In contrast, a style factor consists of all information independent of the identity factor and it does not consider alignment to any single information, hence the entanglement in the t-SNE plot.
\subsubsection{Representation Similarity Analysis}
\begin{figure}
	\centering
	\includegraphics[width=0.99\linewidth]{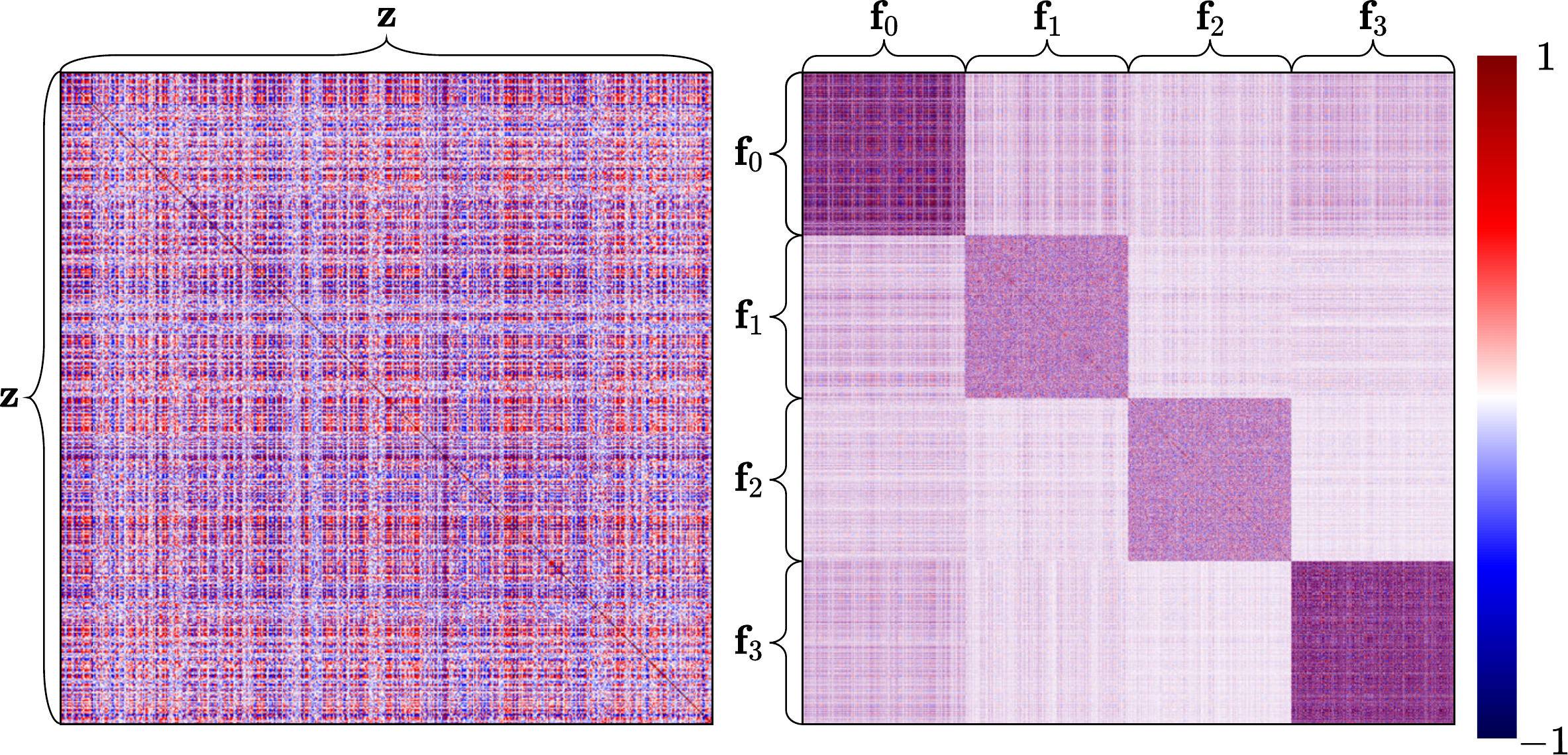}
	\caption{Representational Similarity Analysis (RSA) on units of representation $\textbf{z}$ and units of four factors from Pioneer Network trained on CelebA-128 dataset.
		Values close to 0 are dissimilar, whereas values away from 0 are similar.
		There is a high correlation among units within a factor and significantly low correlation among units of other factors, suggesting that factors do indeed show independence from one another.
		(best viewed in color).
	}
	\label{correlation}
\end{figure}

Representation Similarity Analysis~(RSA)~\cite{kriegeskorte2008representational} is a data analysis framework for comparing dissimilarity between two random variables.
We have drawn a dissimilarity matrix by computing Pearson's correlation coefficient ($r$) for each unit of all factors and each unit of representation $\textbf{z}$ against all other units (Fig.~\ref{correlation}).
As for the RAS on the units of representation $\textbf{z}$, we see high similarity among each units.
However, there is a high correlation among units within a factor and very low correlation among units of other factors, suggesting that factors do indeed show independence from one another.

\subsection{Ablation Study}
\subsubsection{Without Gradient Reversal Layer}
\label{grl_ablate}
\begin{figure}
	\centering
	\includegraphics[width=0.9\linewidth]{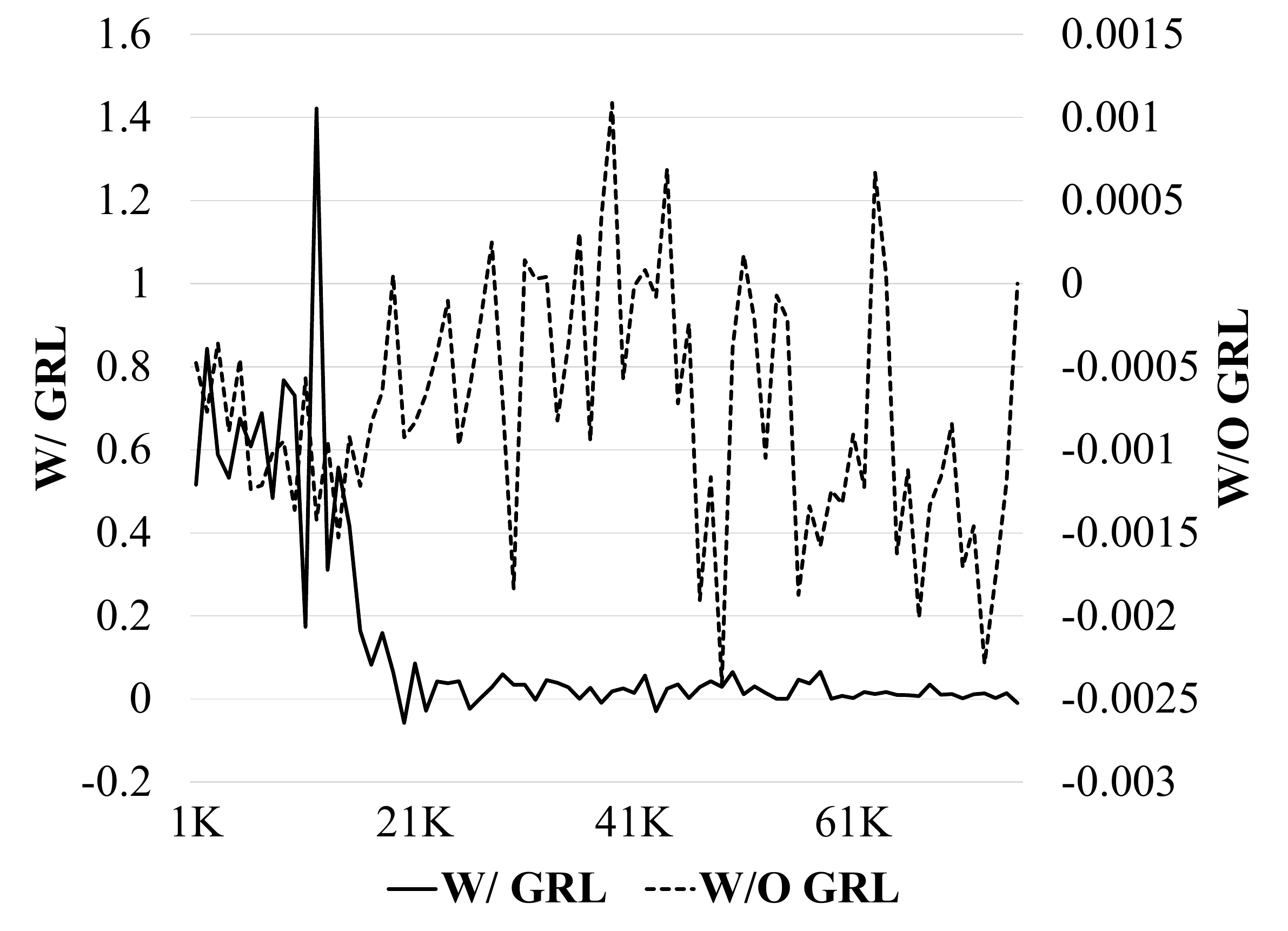}
	\caption{Mutual information training curve with and without GRL. Statistician Network without GRL minimizes the mutual information by first pretraining it with $-\mathcal{L}_M$ and fine-tuning it with~$+\mathcal{L}_M$.
	}
	\label{grl_figure}
\end{figure}

First, we start by replacing the GRL, which is the component responsible for minimizing mutual information (Fig.~\ref{grl_figure}).
To minimize the mutual information without GRL, we pretrain FDEN with negative $\mathcal{L}_M$ for 20,000 iterations and fine-tune with positive $\mathcal{L}_M$.
The mutual information for the FDEN without GRL is steady around 0 for most of the training iterations, suggesting that mutual information is not estimated properly throughout the training procedure.
In contrast, the mutual information for the FDEN with GRL is very high during the beginning of the training iteration and then reduces down to 0 after 20,000 iterations.
This suggests that FDEN is indeed learning to calculate the mutual information in the first 20,000 iterations, and begins to minimize mutual information after 20,000 iterations.

\begin{figure}
	\centering
	\includegraphics[width=0.99\linewidth]{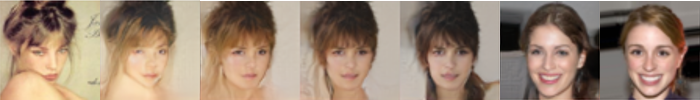}
	\caption{Result of style transfer without Factorizer. The pretrained network is Pioneer Network pretrained on CelebA-64.
		FDEN decomposes the representation into four factors, and the interpolation is performed for one of the factors only (rest of the factors are from left image).
	}
	\label{wo_factorizer}
\end{figure}

\subsubsection{Without Factorizer}
Factorizer is responsible for factorizing a representation into independent and interpretable factors.
Removing the Factorizer from FDEN essentially makes it an autoencoder with multiple streams in the middle.
Although this autoencoder can reconstruct images well, its factors are not independent, nor are they interpretable.
By interpolating only one factor and fixing the other factors (Fig.~\ref{wo_factorizer}), we can see multiple factor of variations, \eg, hair, lips, rotation.
This is comparable to the FDEN with Factorizer (Fig.~\ref{i2i_trans}) that can interpolate factors separately.

\section{Discussion}
\subsection{Low Quality Pretrained Networks}
\begin{figure}
	\centering
	\includegraphics[width=0.85\linewidth]{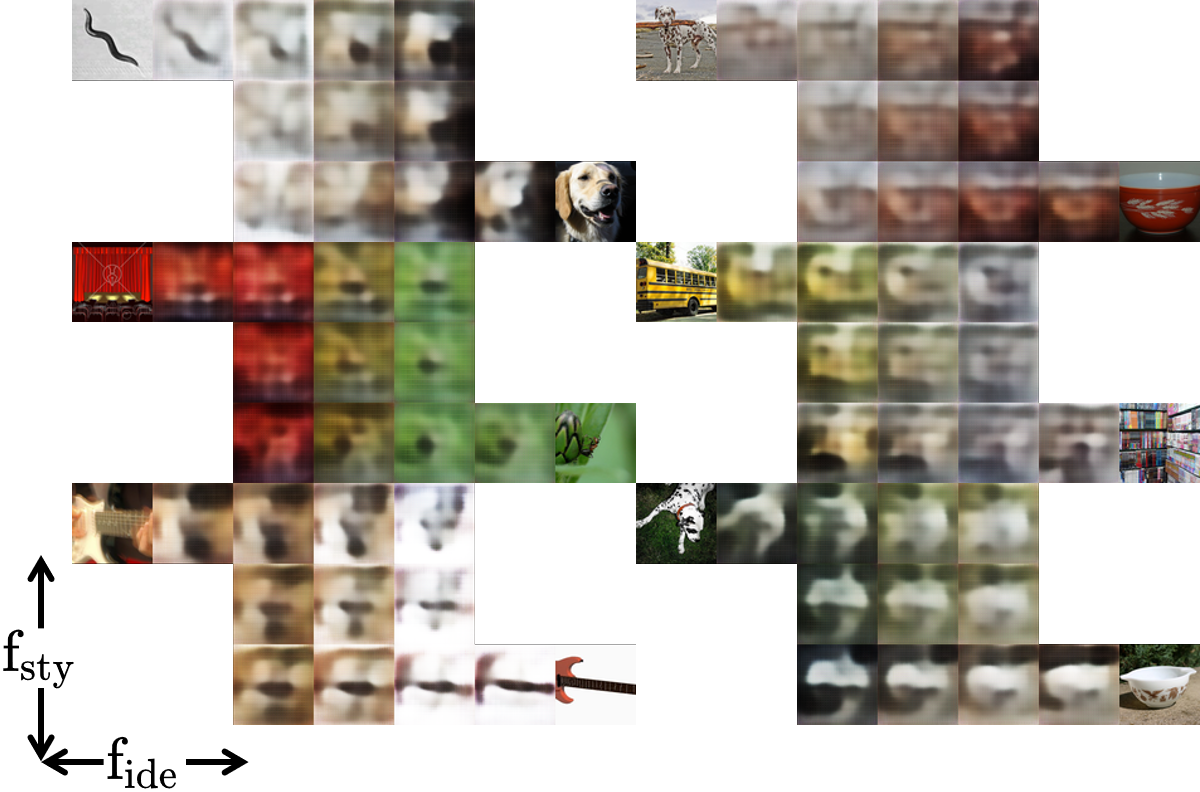}
	\caption{
	Results of image-to-image translation with FDEN with a pretrained ALI.
	The low quality interpolation is due to the low quality reconstruction performance of the pretrained ALI (images on \nth{2} and \nth{6} columns).
	The same weights were used for this experiment as the few-shot learning experiment in Table~\ref{fewshot_acc}, showing that FDEN is able to extract relevant information even with a low-quality pretrained network.
	}
	\label{imagenet_bad_recon}
\end{figure}

{\RJS Since the weights of the pretrained network are fixed while FDEN is trained, the performance of the downstream task is upper bounded by the representative power of the pretrained network (Fig.~S4 in the Supplementary}).
This upper bound is more apparent in image-to-image translation and style transfer because the translated images are combinations of reconstructed images from the pretrained network (\ie, the second and sixth images in Fig.~\ref{i2i_trans}) and not the data samples (\ie, the first and last images in Fig.~\ref{i2i_trans}).
Recent literature have suggested that GANs and autoencoders have a tendency to leave out non-discriminative features during reconstruction~\cite{manisha2018generative}.
To demonstrate this limitation, we applied FDEN to a pretrained autoencoder with low reconstruction performance (\ie, ALI with Mini-ImageNet, Fig.~\ref{imagenet_bad_recon}).
Notably, FDEN could perform image-to-image translation and few-shot learning comparable to other competing methods.

\subsection{Total Correlation}
The DV representation of the KL-divergence requires \textit{i.i.d.} shuffle in the batch axis owing to the marginal distribution, \ie, the latter term in~(\ref{mine_eq}).
With more variables, it becomes difficult to simplify the marginal distribution successfully owing to the shuffling procedure.
In our experiment with on the style transfer, we find that the reconstruction quality is highly correlated with the number of variables and the batch size.
A possible future work for mitigating these limitations is to exploit the representation more closely into the units~\cite{bau2019seeing} rather than factors for a better reconstruction performance.
In doing so, each unit can be considered as a data point to make the shuffling procedure more efficient.
{\RJS Also, some recent works have criticized the KL-Divergence term in mutual information for its large gap in the upper bound~\cite{mcallester2020formal}.
Therefore, a possible future work could be on alleviating this gap (\eg, adding Wasserstein dependency measure~\cite{ozair2019wasserstein}).}
{\RJS \subsection{Decomposition Inconsistency}
\rev{We suspect such} inconsistency in style transfer or image-to-image translation results \rev{with a reason} that the factors are aligned to ``human-labeled'' attributes.
While FDEN tries to produce independent factors, these factors are aligned to factor of variations that are not independent.
For example, there exist dependencies between attributes such as baldness and age (\ie, older people tend to experience baldness more), and beard and gender (\ie, men tend to have beard significantly more than women).
Furthermore, some attributes include a large \rev{within-variations} (\eg, most facial images with bald attribute have receding hair line, while only some images have shaved hair).

Therefore, factor decomposition consistency is related to the classifier performance of the \textit{Alignment Network}, while the decomposition quality is related to the mutual information approximation of the \textit{Statistician Network}.
However, as the classifier performance is upper bounded by the pretrained network, the decomposition inconsistency \rev{could be observable depending on the representations learned in the pretrained network}.
}

\section{Conclusion}
We proposed Factors Decomposer-Entangler Network (FDEN) that learns to decompose a latent representation into independent factors. The results of this study herald the possibility of extending the state-of-the-art models to undertake various tasks without compromising their primary performances.


\ifCLASSOPTIONcaptionsoff
  \newpage
\fi

\bibliographystyle{IEEEtran}
\bibliography{main}

\newpage
\begin{IEEEbiography}[{\includegraphics[width=1in,height=1.25in,clip,keepaspectratio]{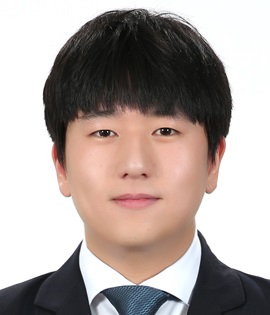}}]{Jee Seok Yoon}
	received the B.S. degree in Computer Science and Engineering from Korea University, Seoul, South Korea, in 2018. He is currently pursuing the Ph.D. degree with the Department of Brain and Cognitive Engineering, Korea University, Seoul, South Korea. 
    
    His current research interests include computer vision, meta learning, and representation learning. 
\end{IEEEbiography}
\enlargethispage{-9.5cm}\vfill
\begin{IEEEbiography}[{\includegraphics[width=1in,height=1.25in,clip,keepaspectratio]{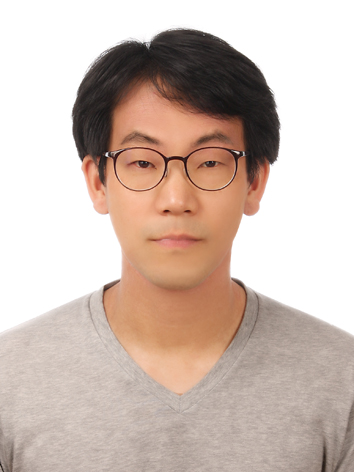}}]{Myung-Cheol Roh} received the Ph.D. degrees in Computer Science and Engineering from Korea University, Seoul, Korea, in 2008.
He worked at the Center for Vision, Speech and Signal Processing in the University of Surrey, UK, as a collaborate researcher in 2004 and at the Robotics Institute in Carnegie Mellon University, US, as a researcher from 2008 to 2012.

From 2012 to 2016, he worked at Samsung S-1, Korea and currently, he is working at Kakao Enterprise, Korea.
His research interests include machine learning, pattern recognition, and face analysis.
\end{IEEEbiography}

\begin{IEEEbiography}[{\includegraphics[width=1in,height=1.25in,clip,keepaspectratio]{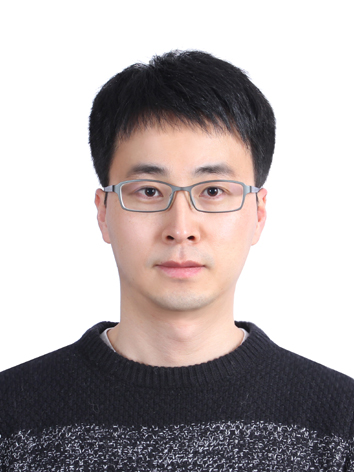}}]{Heung-Il Suk}
	received the Ph.D. degree in computer science and engineering from Korea University, Seoul, South Korea, in 2012.
	
	From 2012 to 2014, he was a Post-Doctoral Research Associate with the University of North Carolina at Chapel Hill, Chapel Hill, NC, USA. He is currently an Associate Professor with the Department of Artificial Intelligence and the Department of Brain and Cognitive Engineering, Korea University. 
    His current research interests include machine learning, biomedical data analysis, brain-computer interface, and healthcare.

    Dr. Suk is serving as an Editorial Board Member for Electronics, Frontiers in Neuroscience, International Journal of Imaging Systems and Technology (IJIST), and a Program Committee or Reviewer for NeurIPS, ICML, ICLR, AAAI, IJCAI, MICCAI, AISTATS, \etc.
\end{IEEEbiography}

\newpage
\onecolumn


\renewcommand\thefigure{S\arabic{figure}}
\renewcommand\thetable{S\arabic{table}}
\setcounter{figure}{0}
\setcounter{table}{0}
\renewcommand\thesection{\thechapter \Alph{section}} 
\renewcommand\thesubsection{\Alph{section}.\arabic{subsection}}

\setcounter{page}{1}
\begin{appendices}
\begin{center}\LARGE\bfseries
Supplementary Material
\end{center}

\section*{Chapter A Additional Results}
\label{add_res}

\subsection{Image-to-image Translation}
We used the images in the first and the last column as the input images for translating. The images in the second and the sixth column are ALI's original reconstruction. The images in the middle are the results of reconstruction with interpolated identity and style factors of the input images.

\begin{figure}[h]
	\centering
	\includegraphics[width=0.74\linewidth]{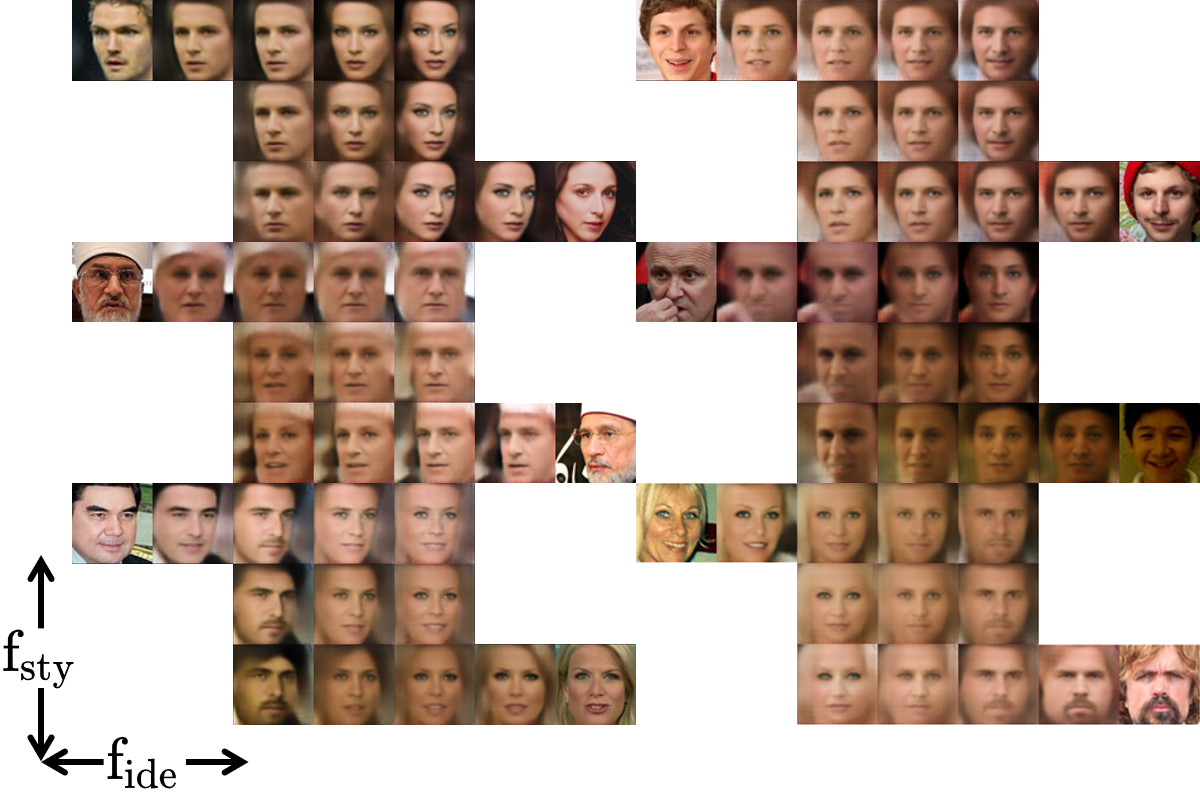}
	\caption{
	Additional results on MS-Celeb-1M data set.
	}
\end{figure}
\begin{figure}[b]
	\centering
	\includegraphics[width=0.74\linewidth]{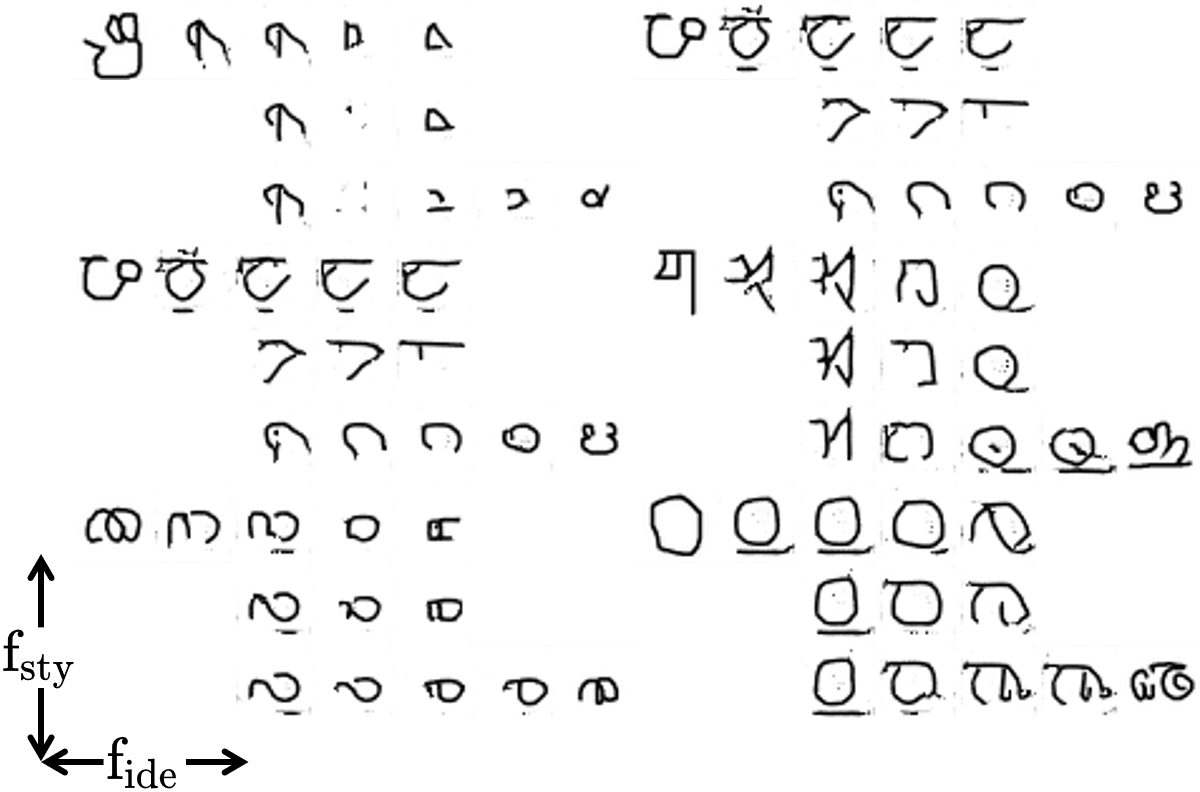}
	\caption{
	Additional results on Omniglot data set.
	}
\end{figure}
\begin{figure}[t]
	\centering
	\includegraphics[width=0.9\linewidth]{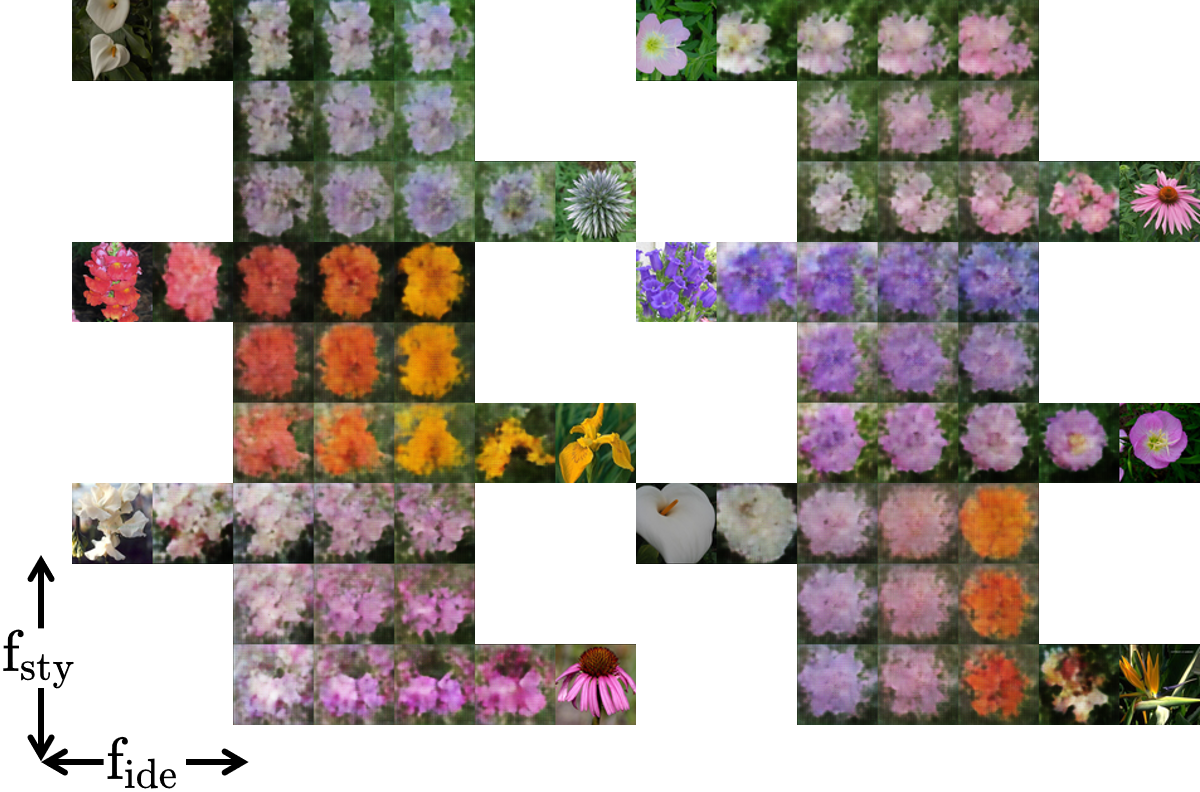}
	\caption{
	Additional results on Oxford Flower data set.
	}
\end{figure}

\subsection{Style Transfer}
\begin{figure}[h]
	\centering
	\includegraphics[width=0.5\linewidth]{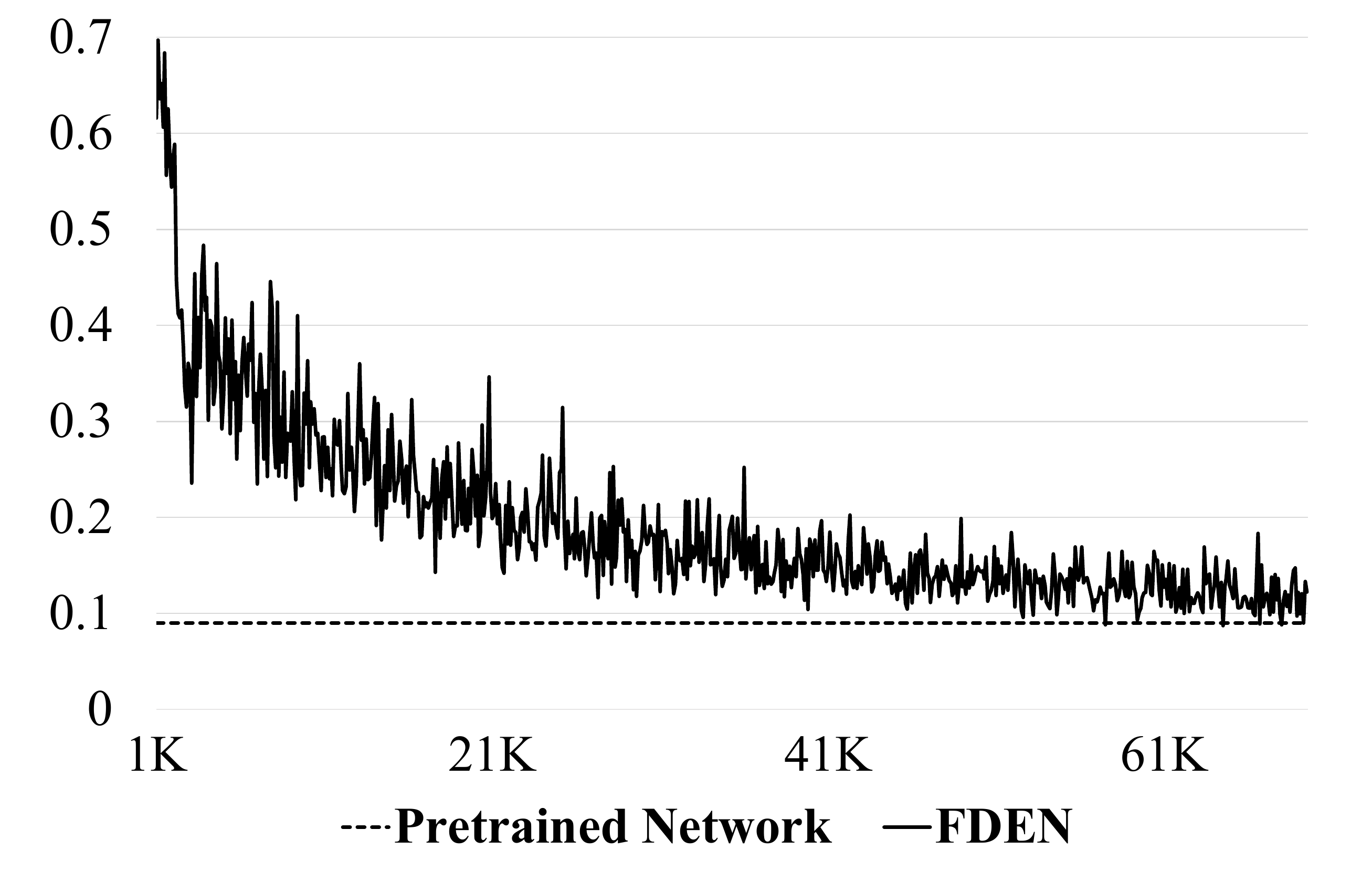}
	\caption{{\RJS Pixel-wise reconstruction loss curve (\ie, $\lvert\lvert \textbf{x} - \tilde{\textbf{x}} \rvert\rvert^{2}_{2}$). The dotted line is the reconstruction loss for the Pioneer Network, and the solid line is for FDEN with CelebA-128 on a style-transfer downstream task. This shows that FDEN is able to reconstruct images with similar quality as the pretrained network with the additional ability to perform style transfer downstream task.}}
	\label{recon_curve}
\end{figure}
\clearpage
\section*{Chapter B Hyperparameters}
\label{hyperpara}
\subsection{FDEN}
\begin{table}[h]
\renewcommand{\arraystretch}{1.127}
\centering
\caption{Model hyperparameters.}
\label{tab:omni-hyper}
\begin{tabular}{@{}rlccl@{}}
\toprule
Operation & Feature Maps & Batch Norm & Dropout & Activation \\ \midrule
$\mathcal{D}_{\theta_{\text{dec}}}\left(\textbf{z}\right) - Dim$ input &  &  &  &  \\
Fully Connected & 512 & $\surd$ & 0.2 & Leaky ReLu \\
Fully Connected & 512 & $\times$ & 0.2 & Leaky ReLu \\
Fully Connected & 512 & $\times$ & 0.2 & Leaky ReLu \\
Fully Connected & $Dim\times 2$ & $\times$ & 0.2 & Linear \\
$\mathcal{D}_{\theta_{\text{i}}}\left(\textbf{z}_{\text{dec}}\right) \forall i \in N- Dim\times 2$ input &  &  &  &  \\
Fully Connected & 512 & $\surd$ & 0.2 & Leaky ReLu \\
Fully Connected & 512 & $\times$ & 0.2 & Leaky ReLu \\
Fully Connected & $Dim$ & $\times$ & 0.2 & Linear \\
$\mathcal{F}_{\psi_{\text{i}}}\left(\textbf{f}_{i}\right) \forall i \in N- Dim$ input &  &  &  &  \\
Fully Connected & 512 & $\surd$ & 0.2 & Leaky ReLu \\
Fully Connected & 256 & $\times$ & 0.2 & Leaky ReLu \\
Fully Connected & 64 & $\times$ & 0.2 & Leaky ReLu \\
Fully Connected & 1 & $\times$ & 0.2 & Linear \\
$\mathcal{F}_{\xi_{\text{mi}}}\left(\textbf{f}_{0},...,\textbf{f}_{N}\right) - Dim$ input &  &  &  &  \\
\multicolumn{5}{c}{Concatenate $\textbf{f}_{\text{0}},...,\textbf{f}_{N}$ along the channel axis} \\
Fully Connected & 1024 & $\surd$ & 0.2 & Leaky ReLu \\
Fully Connected & 256 & $\times$ & 0.2 & Leaky ReLu \\
Fully Connected & 64 & $\times$ & 0.2 & Leaky ReLu \\
Fully Connected & 1 & $\times$ & 0.2 & Linear \\ \midrule[0.1pt]
$\mathcal{E}_{\phi_{i}}\left(\textbf{f}_{i}\right) \forall i\in N - Dim$ input &  &  &  &  \\
Fully Connected & 256 & $\surd$ & 0.2 & Leaky ReLu \\
Fully Connected & 256 & $\times$ & 0.2 & Leaky ReLu \\
Fully Connected & $Dim$ & $\times$ & 0.2 & Linear \\
$\mathcal{E}_{\phi_{\text{enc}}}\left(\tilde{\textbf{f}}_{0},...,\tilde{\textbf{f}}_{N}\right) - Dim$ input &  &  &  &  \\
\multicolumn{5}{c}{Concatenate $\tilde{\textbf{f}}_{0},...,\tilde{\textbf{f}}_{N}$ along the channel axis} \\
Fully Connected & 512 & $\surd$ & 0.2 & Leaky ReLu \\
Fully Connected & 512 & $\times$ & 0.2 & Leaky ReLu \\
Fully Connected & 512 & $\times$ & 0.2 & Leaky ReLu \\
Fully Connected & $Dim$ & $\times$ & 0.2 & Linear \\ \midrule
Optimizer & \multicolumn{4}{l}{Adam $\left(\eta=0.0001, \beta_1=0.5, \beta_2=0.999\right)$} \\
Batch size & \multicolumn{4}{l}{16} \\
Episodes per epoch & \multicolumn{4}{l}{10,000} \\
Epochs & \multicolumn{4}{l}{1,000} \\
Leaky ReLu slope & \multicolumn{4}{l}{0.01} \\
Weight initialization & \multicolumn{4}{l}{Truncated Normal ($\mu=0,\sigma=0.001$)} \\
Loss weights & \multicolumn{4}{l}{$\alpha=1,\beta=1,\gamma=0.5, \lambda=0.5$} \\
$Dim$ & \multicolumn{4}{l}{\begin{tabular}[c]{@{}l@{}}Omniglot - 256\\ MS-Celeb-1M, Mini-ImageNet, Oxford, CelebA - 512\end{tabular}} \\ \bottomrule
\end{tabular}
\end{table}

\subsection{Adversarially Learned Inference}
We chose ALI~\cite{dumoulin2017} for the invertible network of our framework.
We used the exactly the same hyperparameters presented in Chapter A in~\cite{dumoulin2017}.
For training Omniglot data set, we used the model designed for unsupervised learning of SVHN.
For training Mini-ImageNet, MS-Celeb-1M, Oxford Flower data sets, we used the model designed for unsupervised learning of CelebA.
Although~\cite{dumoulin2017} designed a model for a variation of ImageNet (Tiny ImageNet), our preliminary results showed that CelebA model could synthesize better images with Mini-ImageNet data set.

For training Mini-ImageNet, MS-Celeb-1M, Oxford Flowers data sets, we've included a $\ell2$ reconstruction loss between the input image and its reconstructed image. This results in steady convergence and better reconstruction.
\subsection{Pioneer Network}
We chose Pioneer Network~\cite{heljakka2018pioneer} for its state-of-the-art reconstruction performance. We use the pre-trained model for CelebA-128 publicly open at author’s website.
\end{appendices}

\end{document}